\documentclass[10pt,twocolumn,letterpaper]{article}

\usepackage{iccv}
\usepackage{times}
\usepackage{epsfig}
\usepackage{graphicx}
\usepackage{amsmath}
\usepackage{amssymb}

\usepackage[ruled]{algorithm2e}
\SetAlCapNameFnt{\footnotesize}
\SetAlCapFnt{\footnotesize}
\SetAlgorithmName{Algorithm}{Algo}{List of Algorithms}

\usepackage[export]{adjustbox} 
\usepackage{comment}
\usepackage{graphicx}
\usepackage{amsmath}
\usepackage{amssymb}
\usepackage{booktabs}
\usepackage{array}
\newcolumntype{P}[1]{>{\centering\arraybackslash}p{#1}}
\usepackage{multirow}
\usepackage{makecell}
\usepackage{stfloats}
\usepackage{xcolor}
\definecolor{mygray}{HTML}{707070}
\definecolor{mygreen}{rgb}{0.0, 0.5, 0.0}
\usepackage{authblk}
\newcommand*{\affaddr}[1]{#1}
\newcommand*{\affmark}[1][*]{\textsuperscript{#1}}
\newcommand*{\email}[1]{\small\texttt{#1}}
\usepackage[accsupp]{axessibility}  

\usepackage{pifont}
\newcommand{\cmark}{\ding{51}}%
\newcommand{\xmark}{\ding{55}}%
\newcommand{\myfont}[1]{{\fontfamily{qcr}\selectfont {#1}}}

\usepackage[pagebackref=true,breaklinks=true,letterpaper=true,colorlinks,bookmarks=false]{hyperref}

\iccvfinalcopy 

\def\iccvPaperID{1793} 

\ificcvfinal\fi

\begin{document}
\newcommand\abbrev{SINC}
\newcommand\abbrevlong{self-supervised in-context learning}

\title{SINC: Self-Supervised In-Context Learning for Vision-Language Tasks}
\author{
Yi-Syuan Chen\affmark[1], Yun-Zhu Song\affmark[1], Cheng Yu Yeo\affmark[1], Bei Liu\affmark[2], Jianlong Fu\affmark[2], and Hong-Han Shuai\affmark[1]\\\vspace{-10pt}
\affaddr{\affmark[1]National Yang Ming Chiao Tung University, \affmark[2]Microsoft Research Asia} 
\email{\{yschen.ee09,yzsong.ee07,boyyeo123.ee08,hhshuai\}@nycu.edu.tw}\\
\email{\{Bei.Liu,jianf\}@microsoft.com}\\
}


\maketitle
\ificcvfinal\thispagestyle{empty}\fi

\begin{abstract}
Large Pre-trained Transformers exhibit an intriguing capacity for in-context learning. Without gradient updates, these models can rapidly construct new predictors from demonstrations presented in the inputs. Recent works promote this ability in the vision-language domain by incorporating visual information into large language models that can already make in-context predictions. However, these methods could inherit issues in the language domain, such as template sensitivity and hallucination. Also, the scale of these language models raises a significant demand for computations, making learning and operating these models resource-intensive. To this end, we raise a question: ``How can we enable in-context learning without relying on the intrinsic in-context ability of large language models?". To answer it, we propose a succinct and general framework, \textbf{S}elf-supervised \textbf{IN}-\textbf{C}ontext learning (\abbrev), that introduces a meta-model to learn on self-supervised prompts consisting of tailored demonstrations. The learned models can be transferred to downstream tasks for making in-context predictions on-the-fly. Extensive experiments show that \abbrev~outperforms gradient-based methods in various vision-language tasks under few-shot settings. Furthermore, the designs of \abbrev~help us investigate the benefits of in-context learning across different tasks, and the analysis further reveals the essential components for the emergence of in-context learning in the vision-language domain.
\end{abstract}

\section{Introduction}
\begin{figure}[t]
    \centering
    \includegraphics[width=\linewidth]{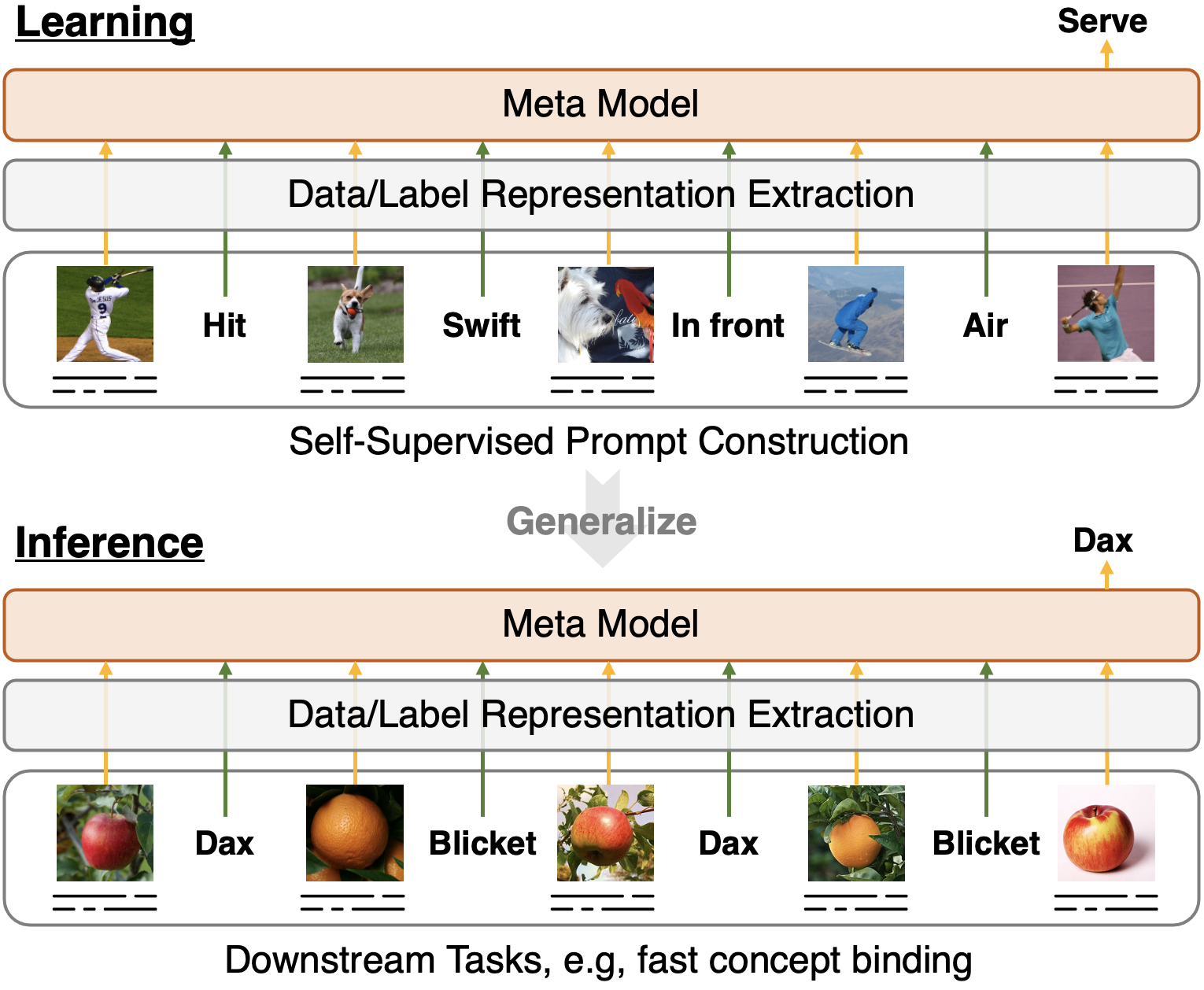}
    \caption{\textbf{Illustration of \abbrev.} A meta-model is introduced for acquiring in-context ability given features extracted from general models. During pre-training, the meta-model is learned on prompts constructed in a self-supervised manner. Our designs jointly enable the transfer of in-context ability to the downstream.}
    \label{fg:concept}
\end{figure}

Large language models such as GPT-3~\cite{NEURIPS2020_1457c0d6} are able to perform \textit{in-context learning (ICL)}: given a prompt consisting of a series of demonstrations and a query data as input, the model can generate the corresponding prediction without any parameter updates. Meanwhile, recent works show that large vision-language (VL) models~\cite{alayrac2022flamingo,NEURIPS2021_01b7575c} can also possess such an ability. Specifically, these models can rapidly incorporate multimodal information with few demonstrations for tackling a variety of downstream tasks, such as image captioning~\cite{chen2015microsoft}, visual question answering~\cite{Goyal_2017_CVPR}, and fast concept binding~\cite{NEURIPS2020_1457c0d6}. Behind the success, the shared scheme of these approaches is to incorporate visual information into large language models via proposed modules. In particular, the in-context ability of these VL models would significantly rely on the language side. As such, the issues in the language domain could be inherited, such as template sensitivity~\cite{lu-etal-2022-fantastically,rubin-etal-2022-learning} and hallucination~\cite{jimenez-gutierrez-etal-2022-thinking}. Previous studies~\cite{garg2022what,alayrac2022flamingo,NEURIPS2020_1457c0d6,shin-etal-2022-effect,NEURIPS2021_01b7575c} also indicate that the in-context ability scales with the model sizes and barely emerges in smaller models~\cite{NEURIPS2020_1457c0d6}. This property requires current methods to be built upon large language models for leveraging in-context demonstrations. Although previous works typically freeze language models for training efficiency, the language models are still involved in the learning process, creating a non-negligible demand for resources to learn or operate these models~\cite{sung2022lst}. Moreover, the length of demonstrations could readily exceed the practical limitation of most Transformer models~\cite{NEURIPS2020_c8512d14}, especially for vision-language data. 

To this end, we pose a challenging research question: ``\textit{How can we enable in-context learning without relying on the intrinsic in-context ability of large language models?}" The access to the solution could lie in the understanding of ICL properties in large language models. Previous studies have shown that the formats of demonstrations can affect performance drastically~\cite{liu-etal-2022-makes,lu-etal-2022-fantastically,zhang-etal-2022-active,pmlr-v139-zhao21c}. Furthermore, \cite{min2022rethinking} shows that randomly assigning labels for the demonstrations could barely decrease the performance, while~\cite{kim2022ground} justifies that the phenomenon is not generalized across tasks. In view of these obscure observations, recent works attempt to study ICL specifically from different perspectives.~\cite{akyurek2022learning,dai2022can} show that Transformers trained from scratch can implicitly implement the gradient descent algorithm in their forward pass, and the number of attention layers~\cite{Zheng_2017_ICCV} could further relate to the equivalent learning steps. \cite{chan2022data, shin-etal-2022-effect}, in another way, study the effects of training data for language models. The results reveal that the emergence of ICL could be attributed to certain language properties such as burstiness~\cite{lambiotte2013burstiness,serrano2009modeling}. Overall, these studies suggest that the in-context ability of language models results from multiple factors. Moreover, \textit{this ability could be incidental since typical language modeling is not intentionally designed based on these factors}. Thus, models could exhibit unexpected behaviors, and the acquisition of in-context ability could be inefficient in terms of model capacity.

\begin{figure}[t]
    \centering
    \includegraphics[width=\linewidth]{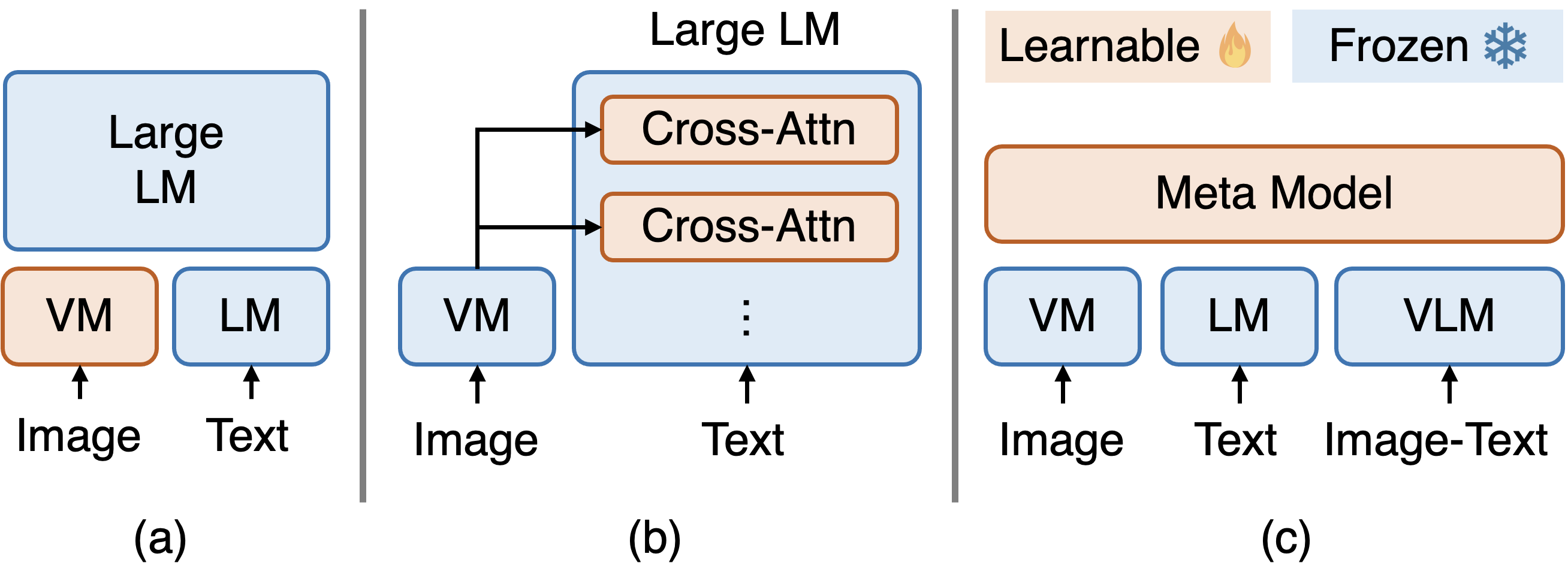}
    \caption{\textbf{Architectural comparison.} Previous works (a)~\cite{NEURIPS2021_01b7575c} and (b)~\cite{alayrac2022flamingo} achieve in-context learning for VL tasks with large language models. Our \abbrev~relieves such a constraint by introducing a meta-model for acquiring the in-context ability.}
    \label{fg:comparison}
\end{figure}

To address previous limitations, we present a general framework, named \textbf{S}elf-supervised \textbf{IN}-\textbf{C}ontext learning (\abbrev). The core idea is to decouple the acquisition of in-context ability from conventional VL pre-training and incentivize it through both architectural and data perspectives. Specifically, we introduce a \textit{meta-model} as the in-context learner that directly operates on the representations produced from frozen models. A self-supervised learning scheme is proposed to enable the meta-model to make predictions based on demonstrations, with the transferability across tasks. In particular, we learn the meta-model on tailored prompts comprising sequences of data and label representations. For constructing data-label pairs in the prompts, inspired by the literature of question answering~\cite{glass-etal-2020-span,ram-etal-2021-shot,xu-lapata-2021-generating}, we regard data sharing similar missing semantics as homogeneous and group them as a class. This strategy enables us to create diverse labels from unannotated image-text pairs. However, we identify that the predictions from models would be agnostic to the demonstrations if there is no adequate correlation with the query data. Therefore, we leverage the idea from few-shot learning~\cite{Chen_Shuai_2021,Sun_2019_CVPR} to create specific prompts to trigger the model for utilizing the demonstrated information. Furthermore, in our observations, downstream tasks would demand the in-context ability to different degrees. We thus propose learning different prompts with a controllable ratio to better benefit and study different tasks. Regarding the formation of representations, on the data side, we propose incorporating pre-trained models from various domains, where the produced features are further aggregated with the proposed \textit{multi-source feature fuser (MFF)}. On the label side, the representations are composed of subword embeddings~\cite{sennrich-etal-2016-neural} of label descriptions, enabling the generalization to unseen labels. Overall, our representation-level in-context learner could be transferred to different scenarios after pre-training, as shown in Fig.~\ref{fg:concept}. A comparison of our architecture with prior works~\cite{alayrac2022flamingo,NEURIPS2021_01b7575c} is depicted in Fig.~\ref{fg:comparison}, wherein prior works either (a) prepend a learnable vision encoder or (b) interleave adapter modules to the large language models for ICL. In contrast, we achieve ICL by introducing the meta-model after the frozen models, enabling us to prepare the representations on separate devices or in an offline manner. This scheme exempts the frozen models from all the backward processes, thereby significantly alleviating the computation burden. The main contributions of this paper are summarized as follows.

\begin{itemize}
    \item We propose a novel framework, SINC, that decouples the acquisition of ICL from VL pre-training, enabling ICL in a more manageable and extensible way  without relying on the intrinsic in-context ability of large language models.
    \item  We propose to learn a meta-model on self-supervised prompts consisting of tailored demonstrations. The learned models can be transferred to downstream tasks for making in-context predictions on-the-fly.
    \item Extensive experiments show that \abbrev~outperforms previous gradient-based methods and a strong ICL baseline. The analysis further reveals the properties and essential components for ICL in the VL domain.
\end{itemize}

\section{Related Works}
\noindent\textbf{Vision-Language Pre-training.} Pre-training for vision-language scenarios is a rapidly evolving domain that aims to bridge the gap between visual perception and natural language comprehension. Present methodologies predominantly employ large-scale transformer-based models, which have showcased impressive effectiveness~\cite{xue2023clipvip,NEURIPS2022_f8290ccc,Xue_2022_CVPR,NEURIPS2021_23fa71cc,Huang_2021_CVPR,huang2020pixel}. Various approaches from distinct perspectives have been proposed for enhancement, such as learning objectives~\cite{Ji_2023_CVPR,Yang_2022_CVPR}, frozen-model utilization~\cite{li2023blip,alayrac2022flamingo}, visual representations~\cite{Huang_2021_CVPR}, alignments~\cite{NEURIPS2021_23fa71cc, xue2023clipvip}, and pre-training datasets~\cite{Xue_2022_CVPR,NEURIPS2021_50525975}. Moreover, a research direction has emerged to further leverage these models in scenarios with limited resources. This research line concentrates on integrating lightweight modules, such as adapters~\cite{Chen_Shuai_2021,9992078}, to enable efficient fine-tuning of vision-language models~\cite{Sung_2022_CVPR, sung2022lst,zhang-etal-2023-hyperpelt}. However, it's important to note that these techniques necessitate alterations to the model architecture, followed by subsequent fine-tuning, which may not be suitable for situations where the model remains inaccessible.

\noindent\textbf{In-Context Learning.}
Since~\cite{NEURIPS2020_1457c0d6} demonstrated the emergence of the in-context learning (ICL) ability in large-scale language models (LLMs), there has been a growing interest in utilizing the ICL paradigm~\cite{holtzman-etal-2021-surface,liu-etal-2022-makes,lu-etal-2022-fantastically,min-etal-2022-noisy,mishra-etal-2022-reframing,razeghi2022impact,rubin-etal-2022-learning,pmlr-v139-zhao21c}. A research line based on pre-trained LLMs has emerged to explain the mechanism of ICL through the lens of pre-training data~\cite{razeghi2022impact,shin-etal-2022-effect,chan2022data}, in-context examples~\cite{min2022rethinking,kim2022ground}, and model architecture~\cite{bansal2022rethinking,olsson2022context}. Studies suggest that the behavior of in-context learners is driven by the distributions of pre-training data, such as burstiness~\cite{lambiotte2013burstiness,chan2022data,serrano2009modeling} and numbers of rarely occurring classes~\cite{razeghi2022impact}. Combining different training corpora can also facilitate the emergence of ICL~\cite{shin-etal-2022-effect}. Further research has found that the label space and input text distribution are more crucial than providing correct labels for demonstrations~\cite{min2022rethinking}, while~\cite{kim2022ground} justifies that the observations could be limited in specific tasks. Additionally, induction heads in large Transformer models may contribute to ICL~\cite{olsson2022context}, and only a few nucleus layers are essential across downstream tasks, suggesting that LLMs may be under-trained~\cite{bansal2022rethinking}. To incorporate ICL into a vision-language model (VLM)~\cite{xue2023clipvip,NEURIPS2022_f8290ccc,Xue_2022_CVPR,NEURIPS2021_23fa71cc,Huang_2021_CVPR,huang2020pixel}, \cite{NEURIPS2021_01b7575c} proposes encoding images into the word embedding space of an LLM, while ~\cite{alayrac2022flamingo} achieves this by interleaving proposed modules to an LLM. However, the potential pitfalls of LLM-based ICL is that LM pre-training is not explicitly designed for this task, and the ability of ICL is therefore implicitly learned as a by-product. Thus, further warmup, calibration, or template designs are usually required~\cite{chen-etal-2022-improving,dong2022survey,holtzman-etal-2021-surface,liu-etal-2022-makes,rubin-etal-2022-learning,pmlr-v139-zhao21c}. Specifically, to bridge the gap between LM pre-training and downstream tasks,~\cite{chen-etal-2022-meta,min-etal-2022-metaicl} apply meta-training and~\cite{kim-etal-2021-changes} explores the prompt-based tuning. These studies indicate that LLMs are not the sole approach for obtaining ICL ability, and a thorough comprehension of underlying factors is critical. Consequently, aside from the LLM-based ICL mentioned above, some works focus on investigating the empirical properties of ICL, such as task construction or model architecture. For instance,~\cite{garg2022what} shows that Transformer models trained from scratch can in-context learn the class of linear functions. \cite{von2022transformers} finds that training Transformers on auto-regressive tasks is closely related to gradient-based meta-learning formulations. \cite{akyurek2022learning} shows that Transformer-based in-context learners implement standard learning algorithms implicitly for linear regressions. Overall, these studies investigate the ICL ability from different perspectives of LLMs to enhance understanding, but most are limited to crafted datasets or simplified architectures. Thus, building on these efforts, we aspire to expand ICL research to more realistic scenarios.

\noindent\textbf{Multimodal Few-shot Learning.}
Few-shot learning has gained prominence in recent years with the rise of pre-trained language models~\cite{10.1145/3582688}. To extend this capability to a multimodal setting, some prior works incorporate lightweight modules, such as adapters~\cite{Chen_Shuai_2021,9992078}, to enable efficient fine-tuning~\cite{Sung_2022_CVPR, sung2022lst,zhang-etal-2023-hyperpelt} with limited data. On the other hand, ~\cite{NEURIPS2021_01b7575c} prepends a trainable vision encoder to a frozen GPT-like LM with 7 billion parameters for ICL. Similarly,~\cite{alayrac2022flamingo} interleaves trainable adapter modules to a frozen LM of 70B parameters and uses in-context examples as prompts. To leverage external knowledge for few-shot learning with GPT-3~\cite{NEURIPS2020_1457c0d6},~\cite{yang2021empirical} converts images to captions to utilize textual demonstrations for ICL. However, the use of large pre-trained VL models could be impractical for real-world applications due to their size. To this end,~\cite{jin-etal-2022-good} examines the effect of prompts and pre-training objectives on relatively smaller few-shot learners. Notably, previous works mainly focus on question-answering or generative tasks, neglecting other reasoning tasks~\cite{suhr-etal-2019-corpus,xie2019visual}.\footnote{\cite{song-etal-2022-clip} proposes a zero-shot approach for SNLI-VE. However, the method is closely tied to additional annotations, as detailed in Appendix.} These tasks are challenging due to their complex semantics, making adaptation from few examples difficult. Thus, we further include these tasks to extensively evaluate our methods.


\section{Methodology}
\begin{figure*}[t]
    \centering
    \includegraphics[width=\textwidth]{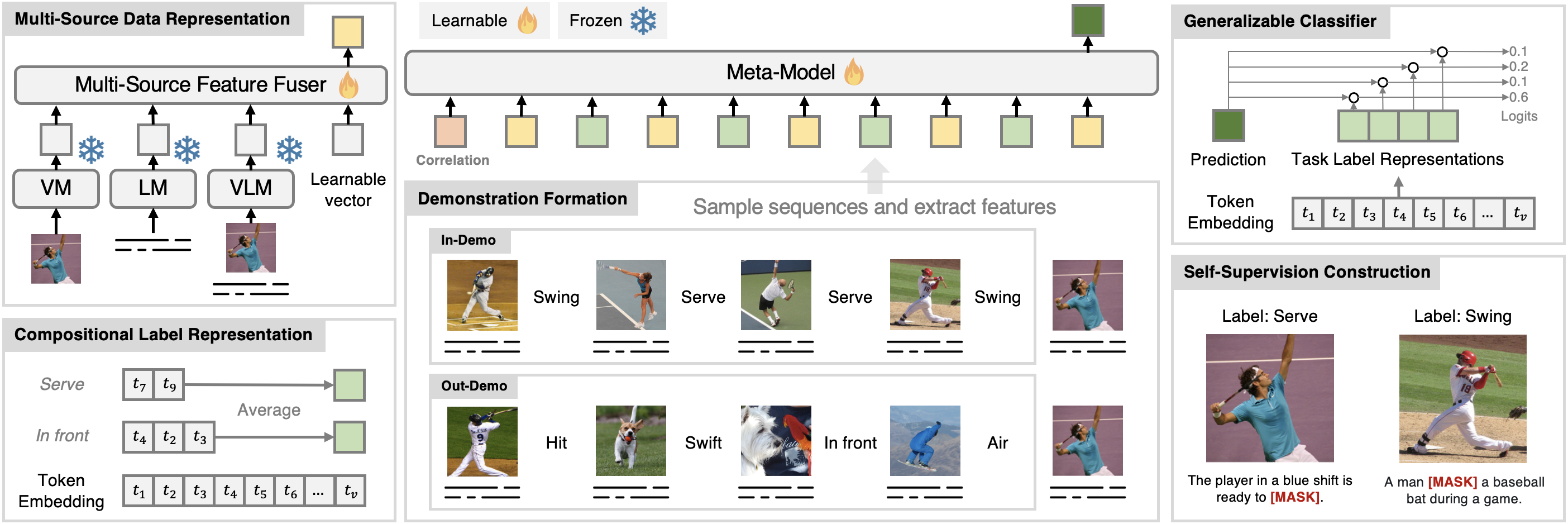}
    \caption{\textbf{Framework of \abbrev.} The meta-model learns on prompts comprising a sequence of data and label representations. Data representation is extracted from multiple pre-trained models (top-left). Label representation is assembled compositionally through pre-trained token embeddings (bottom-left). To incentivize the in-context ability, we construct prompts with tailored demonstrations (bottom-center) in a self-supervised manner (bottom-right). Prediction is then conducted by a classifier that could generalize across tasks (top-right).}
    \label{fg:framework}
\end{figure*}

In this section, we introduce the~\abbrevlong~(\abbrev). We first describe the formulation of our methods, and then introduce the overall framework Finally, we delineate the crucial details of developing the in-context ability in a self-supervised manner based on our framework. The overview of \abbrev~is shown in Fig.~\ref{fg:framework}.


\subsection{Formulation}
\label{subsec:formulation}
Let $\mathcal{F}$ be a set of vision-language tasks and $f \in \mathcal{F}: \mathcal{X}_f \rightarrow \mathcal{C}_f$ is a mapping function, where $\mathcal{X}_f$ is the set of input data and $\mathcal{C}_f$ is the set of classes. A prompt $\pi$ on task $f$ is a sequence $(x^d_1, f(x^d_1),...,x^d_n,f(x^d_n),x^q)$ consisting of a series of demonstrations $\{x^d_i\} \subset \mathcal{X}_f$ and a query data $x^q \in \mathcal{X}_f$. Consider a model $\theta$, we say the model can in-context learn up to $\epsilon$ if it can predict $f(x^q)$ as:
\begin{equation}
\label{eq:in_context_definition}
    \mathbb{E}_{\mathcal{F},\mathcal{X}_f}[\mathcal{L}(\theta(\pi), f(x^q))-\mathcal{L}(\theta(x^q), f(x^q))] \leq \epsilon,
\end{equation}
where $\mathcal{L}$ is an appropriate loss function depending on $f$. Specifically, Eq.~\ref{eq:in_context_definition} indicates whether predictions of a model can be improved with demonstrations. We aim to learn such a model $\theta$ with a pretext task $f^{src}$ and generalize to downstream tasks $\{f_i^{tgt}\} \subset \mathcal{F}$. Furthermore, to investigate the properties of models and tasks, we define the \textit{in-context benefit (ICB)} for a task $f$ on a model $\theta$ as:
\begin{equation}
    ICB(\theta,f) \doteq \mathbb{E}_{\mathcal{X}_f}[
    \frac{\mathcal{L}(\theta(x^q), f(x^q))-\mathcal{L}(\theta(\pi), f(x^q))}{\mathcal{L}(\theta(x^q), f(x^q))}],
\end{equation}
which evaluates the ratio of performance improvements for a model utilizing in-context demonstrations.

\subsection{Overall Framework}
\noindent\textbf{Architecture.} \abbrev~learns a model $\theta$ comprising a \textit{base model} $\theta^{base}$ and a \textit{meta-model} $\theta^{meta}$. Given a prompt $\pi=(x^d_1,f(x^d_1),...,x^q)$, we utilize the base model to extract the representations for the sequence, denoted as $(h^d_1,\widehat{h}^d_1,....,h^q)$. Now, given the representations of preceding demonstrations, we learn the model to predict $f(x^q)$ by minimizing the expected loss:
\begin{gather}
\begin{split}
\label{eq:loss}
    \mathcal{L} & = -\mathbb{E}_{\mathcal{X}_f}[\log P(f(x^q)|\pi,\theta^{base}, \theta^{meta})] \\
    & = -\mathbb{E}_{\mathcal{X}_f}[\log P(f(x^q)|h^q,\{h^d_i\},\{\widehat{h}^d_i\}, \theta^{meta})].
\end{split}
\end{gather}
The meta-model is a decoder-only Transformer~\cite{radford2019language,NEURIPS2020_1457c0d6}, and we only consider the prediction of $f(x^q)$ for loss computation. Next, we address the constructions of $h$ and $\widehat{h}$.

\noindent\textbf{Multi-Source Data Representations.} We design our base model $\theta^{base}$ to flexibly cooperate with multiple knowledge sources. Specifically, given the vision-language data $x$ and several pre-trained models $\{\phi_i\}$ specialized in different modalities, e.g., vision models, language models, and vision-language models, we first extract data features from each of the models with respect to the corresponding modalities, denoted by $\{z_i\}$. Next, we propose a \textit{multi-source feature fuser (MFF)} $\phi^{mff}$ and a learnable indicator $z'$ to aggregate the multimodal information from different sources. The data representation $h$ is then obtained as follows:
\begin{equation}
    h = h_{z'} = \phi^{mff}(z',\{z_i\}),
\end{equation}
where $\phi^{mff}$ is composed of cross-attention layers~\cite{NIPS2017_3f5ee243} and $h_{z'}$ is the output hidden state of $z'$. Overall, the base model $\theta^{base}$ comprises the pre-trained models $\{\phi_i\}$ and the multi-source feature fuser $\phi^{mff}$. Importantly, we keep $\{\phi_i\}$ frozen throughout the learning, which significantly reduce the computation demands. Moreover, attributed to the design of our architecture, $\{z_i\}$ can be prepared offline, as the pre-trained models only need to be forwarded once and can be exempted from all the backward processes, thus achieving better efficiency than the previous method~\cite{NEURIPS2021_01b7575c}.


\noindent\textbf{Compositional Label Representations.} One viable strategy to construct label representations is to create a learnable embedding for each label from scratch. However, such an approach requires maintaining specific embeddings for different tasks, which could impede the model generalizability in downstream tasks since some labels could be unseen during pre-training. To this end, we propose creating label representations in a \textit{compositional} way. The core idea is to leverage pre-trained token embeddings from the base model $\theta^{base}$. Specifically, consider such a token embedding $E \in \mathbb{R}^{|\mathcal{V}| \times m}$, where $\mathcal{V}$ is the vocabulary and $m$ is the embedding size. We tokenize the descriptions of a label $f(x)$ into a sequence $(t_1,...,t_k)$. The label representation is then constructed by averaging over the token embeddings of the sequence as $\widehat{h} = \frac{1}{k}\sum_{i}{E_{t_i}}$, where $E_{t_i} \in \mathbb{R}^m$ is the embedding of token $t_i$. This design allows the model to generalize to various unseen labels in downstream.

\noindent\textbf{Generalizable Classifier.} Following the idea of label representation construction, the classifier should also be generalizable across different tasks. In this regard, we propose to utilize the label representations to the classification process. Specifically, considering a task $f$ with class set $\mathcal{C}_f$, we build the weight matrices of the classifier from the label embeddings $\{\widehat{h}_c | c \in \mathcal{C}_f \}$. For the outputs $\widetilde{h}$ from the meta-model, the predicted probability $P(c)$ of class $c$ is obtained as:
\begin{gather}
    \widetilde{h}' = \sigma(W_1 \widetilde{h}+ b_1),\ \widehat{h}'_c = \sigma(W_2 \widehat{h}_c +b_2), \\
    s_c = W_3(\widetilde{h}' \odot \widehat{h}'_c) + b_3, \\
    P(c) = \exp(s_c) / \textstyle\sum_{c \in \mathcal{C}_f} {\exp(s_c)},
\end{gather}
where $\{W_i, b_i\}_{i=1}^{3}$ are learnable transformation matrices. The classifier is agnostic to the number and content of the labels, offering a universal interface for transferring from pre-training to downstream tasks.

\begin{figure*}[t]
    \centering
    \includegraphics[width=\linewidth]{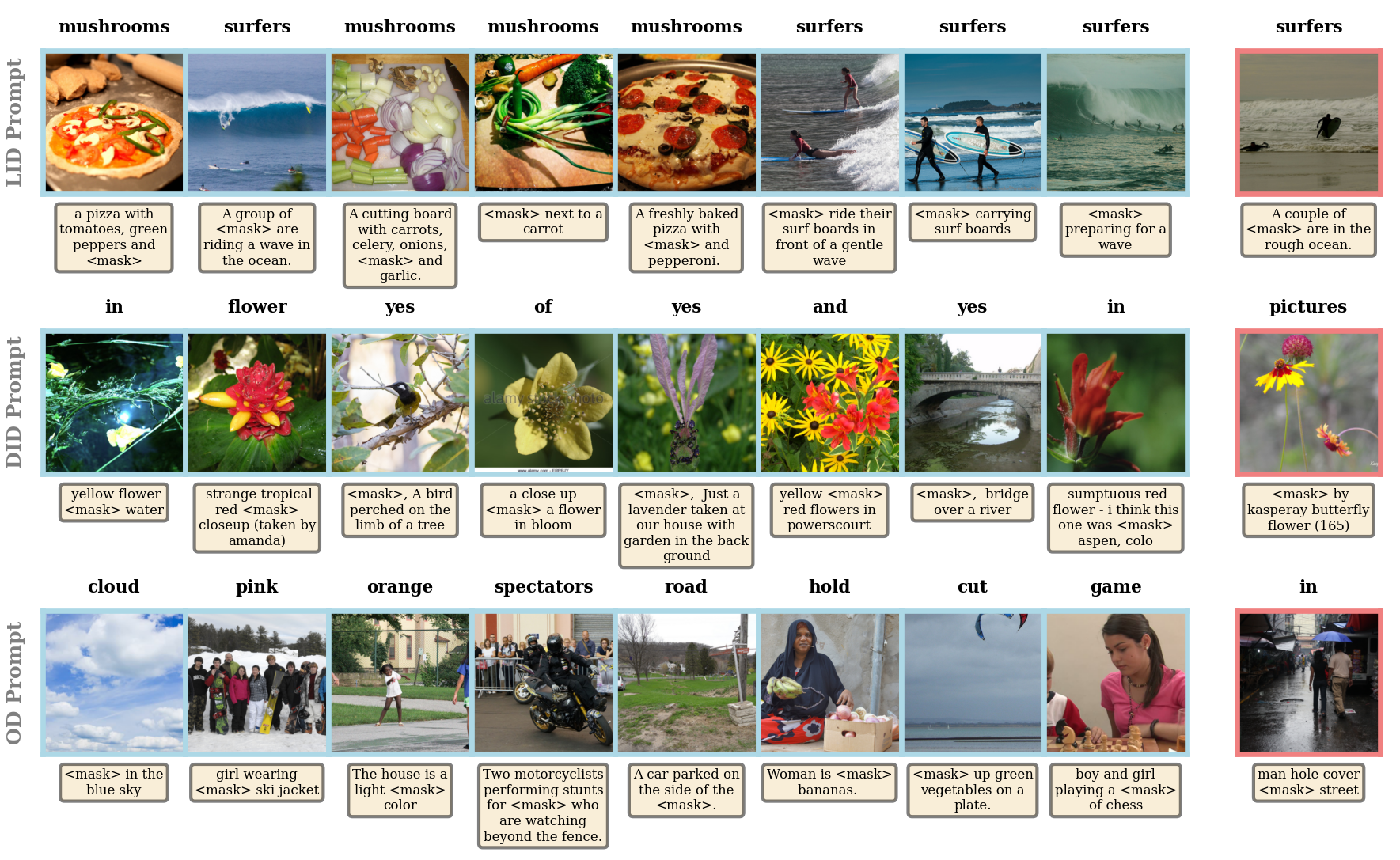}
    \caption{Examples of Label-In-Demo (LID), Data-In-Demo (DID) and Out-Demo (OD) prompts in pre-training. Demonstrations are outlined in blue and query data is outlined in red. Images are center-cropped for better visualization.}
    \label{fg:lid_did_od_prompts}
\end{figure*}

\subsection{Self-Supervised In-Context Learning}
\label{subsec:self_supervised_in_context_learning}
\noindent\textbf{Self-Supervision Construction.} To ensure the generalizability of our framework across various tasks, it is imperative to create prompt sequences that encompass a wide range of data and labels. While collecting data from multiple supervised datasets is a valid approach, the VL domain has limited tasks and label spaces compared to the language~\cite{aribandi2022ext} or vision~\cite{liang2022mvit} domains. To address this issue, we propose creating data-label pairs in a self-supervised manner. Firstly, we establish a general multimodal label set by investigating \textit{concepts} in unannotated image-text pairs. Specifically, we parse texts to identify salient spans that are typically pertinent to both the language and vision domains, such as nouns, verbs, adjectives, and adverbs. The spans are then collected as the pre-training label set $\mathcal{C}$. Next, inspired by the literature of unsupervised question answering~\cite{glass-etal-2020-span,ram-etal-2021-shot,xu-lapata-2021-generating}, we create data for each label by \textit{considering the label as the missing semantics for data}. In particular, for a given label $c \in \mathcal{C}$, we select image-text pairs containing $c$ and replace the span of $c$ with a mask token \myfont{[MASK]}. The mask token indicates missing information, and data queries for the same information can be grouped together. This approach is related to Masked Language Modeling~\cite{devlin-etal-2019-bert}, which aims to predict masked tokens from the context of a sentence, but we concentrate on using the mask token to create homogeneous data for different labels. This design allows us to generate diverse data-label pairs in large quantities for constructing prompts that support generalization.

\noindent\textbf{Demonstration Formation.} Through the utilization of self-supervised data, an extensive amount of prompts can be employed for pre-training. However, we observe that the formation of demonstrations has a significant impact on the emergence of the in-context ability. Specifically, we identify that models tend to disregard the demonstrations and rely solely on the query data for predictions. We believe this is a result of the insufficient correlation between the query data and demonstrations, as models tend to learn from the readily accessible information regarding predictions~\cite{goyal-etal-2022-training,singla2022salient}. To this end, we propose three types of prompts: \textit{label-in-demo (LID)}, \textit{data-in-demo (DID)}, and \textit{out-demo (OD)} prompts, to incentivize the model capacity from various perspectives. The LID and DID prompts aim to enhance the correlation between the query data and demonstrations, whereas the OD prompts aim to reinforce the utilization of query data. Specifically, given a query data $x^q$ belonging to class $c^q$, we sample $n$ classes $\mathcal{C}^d$ from the pre-training label set $\mathcal{C}$, and then sample an equal amount of data for each class to create the LID prompts: $\mathcal{X}^{lid} \subset \{g(c)|c \in (\mathcal{C}^d \cup \{c^q\})\}$, where $g: \mathcal{C} \rightarrow \mathcal{X} $ indicates the set of data with the same class. This approach simulates the few-shot learning regime~\cite{Chen_Shuai_2021,Sun_2019_CVPR} and encourages model learning based on demonstrated information. While LID prompts consider the correlation on the label space, we propose to learn with correlated demonstrations also on the data space. This is achieved by retrieving similar data based on the vision-language representations to create the DID prompts: $\mathcal{X}^{did} \subset top{\text -}k(sim(\mathcal{X}, x^q))$, where $sim(\cdot,\cdot)$ is a simple cosine similarity function. The LID and DID prompt jointly promote the models to utilize demonstrations for predictions. To balance the utilization of query data and demonstrations, the OD prompts are constructed by randomly sampling data from the datasets: $\mathcal{X}^{od} \subset \mathcal{X}$, which has relatively fewer benefits from demonstrations. We further introduce the \textit{in-demo ratio} $\rho$ to balance the exposure of OD or LID/DID prompts during pre-training. This enables us to further control the model's inclination to leverage demonstrations, which could vary across different tasks. The examples of LID/DID/OD prompts are shown in Fig.~\ref{fg:lid_did_od_prompts} and the learning process is summarized in Alg.~\ref{alg:learning_procedure}

\noindent\textbf{Correlation Embeddings.} Utilizing the proposed prompts for learning allows the model to predict with variant reliance on the provided demonstrations. To enable the model to activate the corresponding capacity concerning the demonstrations, we introduce the \textit{correlation embeddings} that specify the relations between the query data and demonstrations. Specifically, we add the embeddings to data representations of prompts, $h^q$ and $\{ \widehat{h}^d_i \}$, as follows:
\begin{gather}
    h^q \leftarrow h^q + e_c,\ h^d_i \leftarrow h^d_i + e_c, \\
    e_c = \sigma(W_c(\textstyle\sum_i{(\widehat{h}^d_i} \odot h^q))+b_c),
\end{gather}
where $W_c$ and $b_c$ are learnable parameters. This design provides controllability of predictions conditioned on demonstrations, benefiting tasks with diverse prompt distributions.

\SetInd{0.5em}{0.5em}
\begin{algorithm}
\footnotesize
\caption{Self-Supervised In-Context Learning}
\label{alg:learning_procedure}
\SetKwInput{KwRq}{Required}

Construct self-supervised dataset $\mathcal{D}=\{(x,f(x))\}$. \\
Let $\mathcal{X}$ be the data set, $\mathcal{C}$ be the class set. \\
\For{$(x^q, f(x^q)) \in \mathcal{D}$}{
    \eIf{$\delta_1 < \rho, \text{where } \delta_1 \sim \mathcal{U}(0,1)$}{
        \eIf{$\delta_2 = 0, \text{where } \delta_2 \sim \mathcal{B}(0.5)$}{
            Sample classes $\mathcal{C}^d \subset \mathcal{C}$\;
            Sample data $\mathcal{X}^d = \mathcal{X}^{lid} \subset \{g(c)|c\in(\mathcal{C}^d \cup \{c^q\})\}$\;
        }{
            Sample data $\mathcal{X}^d = \mathcal{X}^{did} \subset top{\text -}k(sim(\mathcal{X}, x^q))$\;
        }
    }{
        Sample data $\mathcal{X}^d = \mathcal{X}^{out} \subset \mathcal{X}$\;
    }
    $\pi=(\mathcal{X}^{d}, f(\mathcal{X}^{d}), x^q)
        = (x^d_1, f(x^d_1),...,x^q)$\;
    Compute loss $\mathcal{L}(f(x^q)|\pi,\theta)  $ from Eq.~\ref{eq:loss}\ and update $\theta$;
}
\end{algorithm}

\section{Experiments}
\subsection{Experimental Settings}

\noindent\textbf{Pre-training Datasets.}
We construct the self-supervised dataset proposed in Sec.~\ref{subsec:self_supervised_in_context_learning} from four image-text datasets, including COCO~\cite{chen2015microsoft}, Visual Genome~\cite{krishna2017visual}, Conceptual Captions~\cite{sharma-etal-2018-conceptual}, and SBU Captions~\cite{NIPS2011_5dd9db5e}. The labels are designed to encompass nouns, verbs, adjectives, and adverbs extracted from image-text pairs. The dataset is curated to contain 4 million data, and further expansion is feasible. To enhance learning efficiency, we preprocess the data representations offline. 

\noindent\textbf{Downstream Datasets.} We benchmark \abbrev~on various VL tasks, including multimodal fast concept binding~\cite{NEURIPS2021_01b7575c}, visual question answering (VQAv2~\cite{Goyal_2017_CVPR}), visual entailment (SNLI-VE~\cite{xie2019visual}) and visual reasoning ($\text{NLVR}^2$~\cite{suhr-etal-2019-corpus}). These tasks exhibit diverse data formats, enabling us to examine the properties of ICL across different scenarios.

\noindent\textbf{Implementation Details.} The meta-model is a 12-layer decoder-only transformer, and the multi-source feature fuser comprises a single cross-attention layer. For data representation, METER~\cite{Dou_2022_CVPR}, ViT~\cite{dosovitskiy2021an}, and RoBERTa~\cite{liu2019roberta} are considered as the vision-language, vision, and language knowledge sources. During pre-training, we use 8 demonstrations. For DID prompts, we leverage Faiss~\cite{johnson2019billion} to retrieve related data based on VL representations. For LID prompts, we sample 1 class in addition to the query class. The model is pre-trained for 500k steps with 4k warm-up steps. We monitor pre-training performance with LID and OD prompts from a separate validation set. For downstream tasks, we leverage DID prompts for ICL evaluation.


\subsection{Comparison with Prior Arts}
We first conduct experiments on the fast concept binding~\cite{NEURIPS2021_01b7575c}, which is established to evaluate models' ability to associate a word with a visual category in few-shot settings. Tab.~\ref{tab:fast_concept_binding} demonstrates that \abbrev~significantly outperforms both ICL (row 3) and gradient-based (GD) methods (rows 1-2) by at least 57.1\%/50.1\%/51.5\% under 2-/6-/10-shots. The main benefits of Frozen (row 3) come from the huge pre-trained LM, which also limits the model's capacity to adapt to new concepts from a few demonstrations since it depends mostly on pre-learned knowledge. In contrast, \abbrev~possesses the advantage of reducing the dependency on linguistic cues or template designs for leveraging demonstrations. This property allows \abbrev~to better adapt to novel tasks, thereby highlighting its superiority. Notably, GD methods (rows 1-2) hardly learn an effective predictor to tackle novel words, even METER-P reuses the language model head as in the prompt learning scheme~\cite{10.1145/3560815}.

Next, we evaluate \abbrev~on various real-world VL tasks. Tab.~\ref{tab:nlvr_snli} presents the comparisons on visual entailment and reasoning tasks under few-shot regimes, which have received limited attention in prior research. The results demonstrate that \abbrev~is capable of tackling tasks that require reasoning skills. Notably, the data representations for $\text{NLVR}^2$ are obtained by combining two images, which are generally not seen during pre-training, highlighting \abbrev's ability to generalize to diverse data representations. Tab.~\ref{tab:vqa} further presents the comparisons on visual question answering, indicating that \abbrev~outperforms GD methods (rows 4-7) under the 4-shot setting and remains competitive for higher shot numbers. Notably,~\cite{song-etal-2022-clip} (row 7) is tailored for the VQAv2 dataset. It uses a pre-trained language model to filter candidate answers and generate text templates for matching images with CLIP~\cite{pmlr-v139-radford21a}, and the computation cost would increase rapidly with the number of candidate answers. We emphasize that \abbrev~confers further benefits in enabling predictions without any parameter updates. Additionally, our framework is generalizable across different tasks, thus eliminating the need for problem-specific tailoring as prior works. From the comparisons, we also note that the benefits from increasing shot number tend to plateau for \abbrev, which aligns with prior research~\cite{alayrac2022flamingo}. Further investigation is presented in Sec.~\ref{subsec:main_properties}. Compared to previous ICL methods (rows 1-3), \abbrev~employs a significantly lower number of learnable and frozen parameters (at least 3.5 and 13.2 times less) to achieve ICL. While we aim to compare models with a more manageable size ($<$1B), \abbrev~can still outperform Frozen ($>$7B) substantially, emphasizing the merits of learning the ICL explicitly.


\begin{table}[t]
\renewcommand{\arraystretch}{1.0}
\footnotesize
\begin{center}
\begin{tabular}{l|c|c|c}
\toprule
\multirow{2}{*}{Model} & \multirow{2}{*}{GD} & \# of Params. & Fast Concept Binding \\
& & Learn / Frozen & 2- / 6- / 10-shot \\
\midrule
METER-C~\cite{Dou_2022_CVPR}
    & \cmark & 319M / \ \ \ \ - \ \ \ \ & 50.00 / 50.43 / 50.98 \\
METER-P~\cite{Dou_2022_CVPR}
    & \cmark & 319M / \ \ \ \ - \ \ \ \ & 50.00 / 50.33 / 51.20 \\
Frozen~\cite{NEURIPS2021_01b7575c}
    & \xmark & 438M / \ \ 7B \ \ \ & 53.40 / 57.90 / 58.90 \\
\midrule
\multirow{2}{*}{\abbrev (ours)} & \multirow{2}{*}{\xmark}
    & 82M / 319M & 76.56 / 79.88 / 82.28 \\
    & & 124M / 529M & \textbf{83.88 / 86.92 / 89.24} \\
\bottomrule
\end{tabular}
\end{center}
\caption{Performance comparisons on fast concept binding. The "-C" and "-P" methods refer to initializing classifiers or reusing LM heads for predictions. GD specifies the need of gradient descent.}
\label{tab:fast_concept_binding}
\end{table}

\begin{table}[t]
\renewcommand{\arraystretch}{1.0}
\footnotesize
\begin{center}
\begin{tabular}{l|P{0.05\linewidth}|c|P{0.14\linewidth}c}
\toprule
\multirow{2}{*}{Model} & \multirow{2}{*}{GD} & \# of Params. & SNLI-VE & $\text{NLVR}^2$ \\
& & Learn / Frozen & dev / test & dev / test \\
\midrule
VL-T5~\cite{pmlr-v139-cho21a} & \cmark & 224M / \ \ \ \ - \ \ \ \
    & 37.54/38.51 & 52.65/51.50 \\
METER~\cite{Dou_2022_CVPR} & \cmark & 319M / \ \ \ \ - \ \ \ \
    & 49.21/49.14 & 55.03/55.03 \\
\midrule
\multirow{2}{*}{\abbrev (ours)} & \multirow{2}{*}{\xmark} 
    & 82M / 319M & 53.03/53.02 & 55.27/55.30 \\
    & & 124M / 529M & \textbf{54.71/54.98} & \textbf{58.94/59.04} \\
\bottomrule
\end{tabular}
\end{center}
\caption{Performance comparisons on SNLI-VE and $\text{NLVR}^2$ with 16 demonstrations. GD specifies the need of gradient descent.}
\label{tab:nlvr_snli}
\end{table}

\begin{table}[t]
\renewcommand{\arraystretch}{1.0}
\footnotesize
\begin{center}
\begin{tabular}{l|c|c|c}
\toprule
\multirow{2}{*}{Model} & \multirow{2}{*}{GD} & \# of Params. & VQAv2 \\
& & Learn / Frozen & 4- / 16- / 32-shot \\

\midrule
\multicolumn{4}{l}{\textcolor{mygray}{\textit{Frozen parameter counts $>$ 1B}}} \\[-0.2ex]
\midrule

\textcolor{mygray}{Frozen~\cite{NEURIPS2021_01b7575c}} & \xmark
    & \textcolor{mygray}{438M / \ \ 7B \ \ \ }
    & \textcolor{mygray}{38.20 / \ \ \ \ - \ \ \ \ / \ \ \ - \ \ \ } \\
\textcolor{mygray}{PICa~\cite{yang2021empirical}} & \xmark
    & \textcolor{mygray}{ \ \ \ - \ \ \ / 175B}
    & \textcolor{mygray}{\ \ \ \ - \ \ \ / 54.30 / \ \ \ - \ \ \ } \\
\textcolor{mygray}{Flamingo~\cite{alayrac2022flamingo} } & \xmark
    & \textcolor{mygray}{10B / 70.5B}
    & \textcolor{mygray}{63.10 / 66.80 / 67.60} \\

\midrule
\multicolumn{4}{l}{\textit{Frozen parameter counts} $<$ \textit{1B}} \\[-0.2ex]
\midrule

METER~\cite{Dou_2022_CVPR} & \cmark & 319M / \ \ \ \ - \ \ \ \
    & 23.53 / 24.43 / 26.89 \\
VL-T5~\cite{pmlr-v139-cho21a} & \cmark & 224M / \ \ \ \ - \ \ \ \
    & \ \ \ \ - \ \ \ \ / 31.80 / \ \ \ \ - \ \ \ \\
FewVLM~\cite{jin-etal-2022-good} & \cmark & 224M / \ \ \ \ - \ \ \ \
    & 45.10 / 48.20 / \ \ \ \ - \ \ \ \\
TAP-C~\cite{song-etal-2022-clip} & \cmark & 0.3M / 229M
    & 45.87 / \textbf{48.89} / \textbf{50.18} \\
\midrule
\multirow{2}{*}{\abbrev (ours)} & \multirow{2}{*}{\xmark}
    & 82M / 319M & \underline{46.21} / 46.60 / 46.82 \\
    & & 124M / 529M & \textbf{47.21} / \underline{48.25} / \underline{48.58} \\
\bottomrule
\end{tabular}
\end{center}
\caption{Performance comparisons on VQAv2. GD specifies the need of gradient descent. The second best scores are \underline{underlined}. }
\label{tab:vqa}
\end{table}

\subsection{Main Properties}
\label{subsec:main_properties}
\begin{figure}[t]
    \centering
    \includegraphics[width=\linewidth]{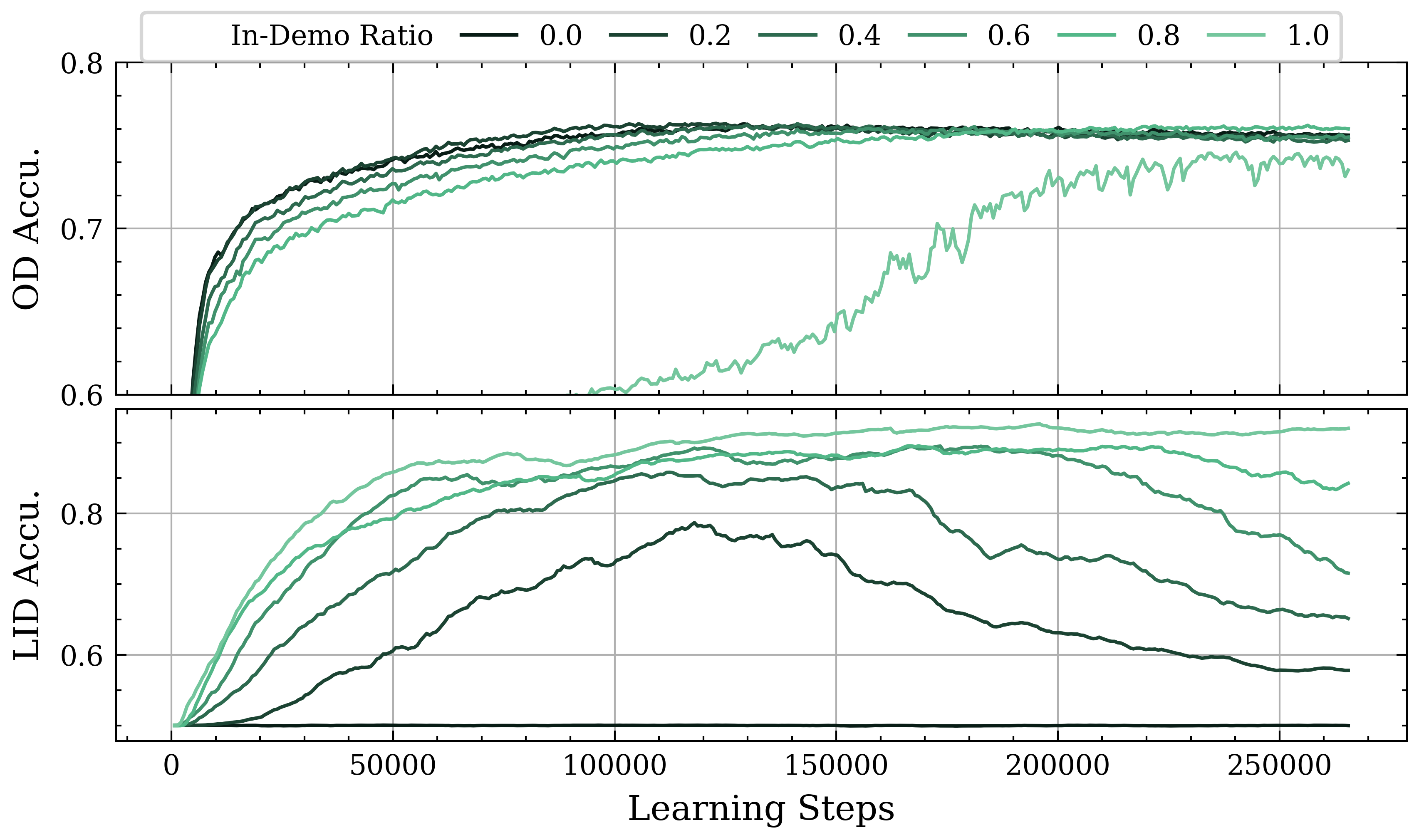}
    \caption{Dynamics of validation performance for OD and LID prompts during pre-training.}
    \label{fg:learning_dynamics}
\end{figure}

\noindent\textbf{Learning Dynamics.} We monitored the performance of the model during pre-training to investigate the emergence of in-context ability in Fig.~\ref{fg:learning_dynamics}. The results suggest a trade-off between the performance of OD and LID prompts in two regards. Firstly, an increase in the in-demo ratio is associated with a higher LID performance and the early development of in-context ability. Notably, during validation, we mapped the labels of the LID prompts to random ones, precluding the possibility of accurate predictions based on memorization, and therefore establishes that the models indeed leverage information from the demonstrations. Secondly, we found that the performance of LID prompts would decrease as the training proceeds to later stages, while the performance of OD prompts continues to improve. We hypothesize that this trade-off could stem from the learning interference across data distributions~\cite{pilault2021conditionally,pmlr-v139-triantafillou21a} since different prompts aim to acquire distinct and possibly opposing capacities. The alleviation of this trade-off is also worth exploring as a future research direction. Furthermore, we observed that the performance of LID prompts remains at binary chances for a 0.0 in-demo ratio. Overall, our findings suggest that models may not efficiently learn to predict with demonstration without specific incentivization, and our framework effectively addresses this issue through the learning with self-supervised prompts.

\begin{figure}[t]
    \centering
    \includegraphics[width=\linewidth]{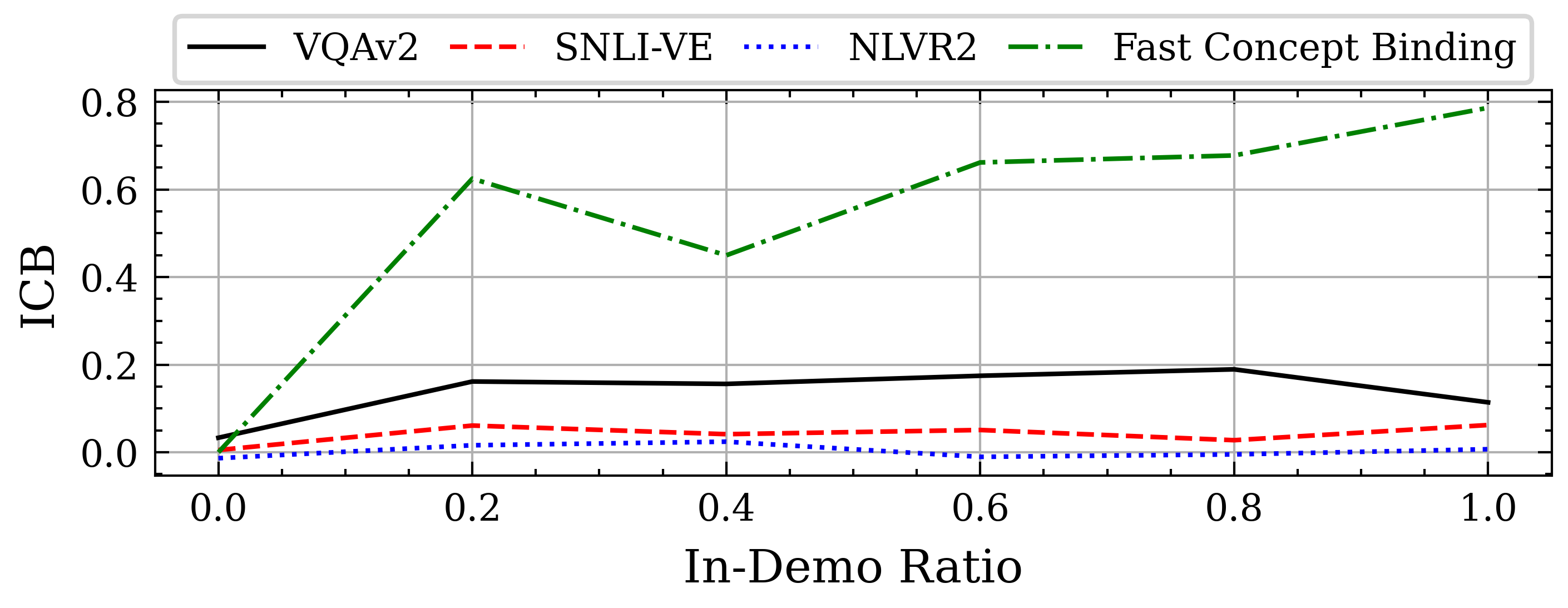}
    \caption{In-Context Benefit (ICB) for models learned in different in-demo ratios. The number of demonstrations is 2 and 4 for fast concept binding and other tasks, respectively.}
    \label{fg:in_demo_ratio}
\end{figure}

\noindent\textbf{Different In-Demo Ratio.}
Fig.~\ref{fg:in_demo_ratio} presents the in-context benefits (ICB) achieved by models trained with different in-demo ratios across various tasks. Our results reveal that a higher in-demo ratio confers significant advantages for tasks such as fast concept binding, which entails simple classification but requires sufficient information from demonstrations. Moreover, we observe that VQAv2 attains a higher ICB compared to $\text{NLVR}^2$ and SNLI-VE. We attribute this finding to their smaller label space, which comprises only two and three categories, thus rendering the construction of the label space relatively easy and relying less on demonstrations. Notably, since the frozen VL model we used~\cite{Dou_2022_CVPR} is pre-trained with image-text matching (ITM), demonstrations may readily activate the binary classification ability, resulting in the closed ICB for $\text{NLVR}^2$ across different in-demo ratios. Importantly, we note that the peak of ICB varies across tasks, indicating that different tasks require varying degrees of in-context ability. We believe this highlights the necessity of providing controllability in leveraging demonstrations, which we primarily achieved through the design of correlation embeddings.

\begin{figure}[t]
    \centering
    \includegraphics[width=\linewidth]{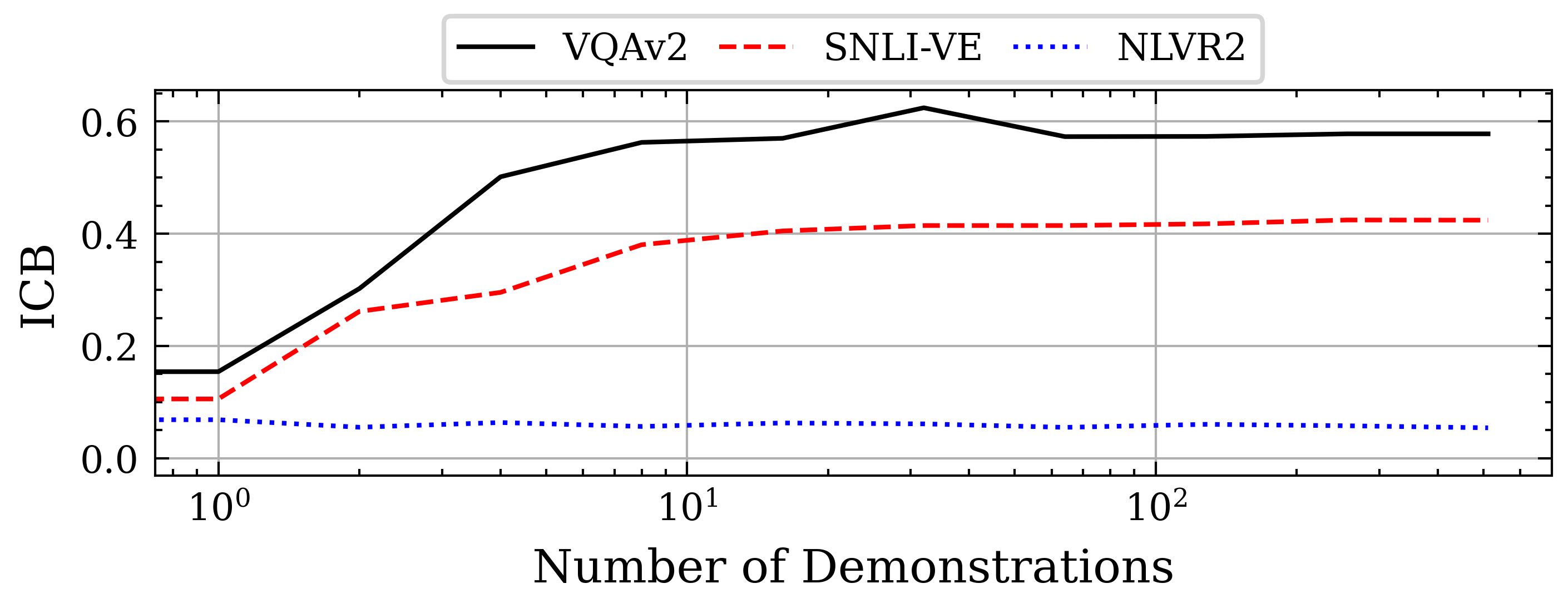}
    \caption{In-Context Benefit (ICB) for different numbers of demonstrations. The model is learned in a 0.2 in-demo ratio.}
    \label{fg:num_of_demos}
\end{figure}

\noindent\textbf{Different Number of Demonstrations.} Fig.~\ref{fg:num_of_demos} shows the impact of varying numbers of demonstrations on the in-context benefit (ICB). Significantly, different tasks exhibit varying sensitivity to the number of demonstrations, with a notable performance saturation as the length of prompts increases. We hypothesize that the saturation may arise due to the generalization issues of Transformer models. Specifically, previous studies have highlighted significant generalization deficiencies in Transformers with respect to sequence length~\cite{anil2022exploring}, and attention could be distracted on long sequences, resulting in degraded performance~\cite{song-etal-2022-improving}. Thus, relevant techniques may be applied to mitigate this issue, which we identify as an important future research direction. Notably, the efficient learning scheme of \abbrev~allows us to investigate such issues with higher shots to further understand the properties of ICB.


\noindent\textbf{Order Sensitivity.} To explore the influence of demonstration order on \abbrev, we evaluate standard deviations for performance on VQAv2/SNLI-VE/NLVR2, yielding respective values of 0.41/0.35/0.53 for the 0.2 in-demo ratio and 4 demonstrations, which is generally not significant compared to mean values. Our results indicate that \abbrev~exhibits low sensitivity to the order of demonstrations. We attribute this property to the specialized learning process of \abbrev, which facilitates exposure to diverse prompts. 

\noindent\textbf{Different Settings.}
We evaluate the efficacy of \abbrev~on various settings, as demonstrated in Tab.~\ref{tab:different_settings}. Our results reveal that decreasing the meta-model size from 124M to 82M does not lead to a significant reduction in performance, while concurrently enhancing operational efficiency (row 2). Furthermore, increasing the size of the pre-training datasets yields improved performance, indicating the potential for further scalability (row 3). Our ablation analysis also demonstrates the performance benefits of utilizing multi-source data representation (rows 4-5). Remarkably, even in the absence of pre-trained VL models, \abbrev~could effectively learn for ICL by leveraging both vision and language representations (row 5). Moreover, our framework is highly flexible, enabling the use of different VL models, e.g., BLIP2~\cite{li2023blip} and ViLT~\cite{pmlr-v139-kim21k}, for learning an ICL framework. The results demonstrate that \abbrev~can generalize across various VL models, with feature quality having a further impact on performance (rows 6-7). Therefore, it allows us to boost performance with stronger feature extraction models. Additionally, we explored the impact of demonstration selection, where OD prompts were used instead of LID prompts for evaluation (row 8). The results indicate that incorporating relevant demonstrations can further boost performance, aligning with prior studies~\cite{liu-etal-2022-makes,zhang-etal-2022-active}.

\subsection{Learning Efficiency}
\abbrev~decouples the learning of in-context ability by introducing a meta-model that operates on the representations generated by frozen models. By doing so, the frozen models do not participate in the backward process, thereby reducing the computational cost significantly. To evaluate the effectiveness of this scheme, we conduct experiments by comparing the learning cost, including GFLOPs and memory footprint of \abbrev~with that of the Frozen~\cite{NEURIPS2021_01b7575c}, under comparable settings. More details are available in Appendix. The results, presented in Tab.~\ref{tab:learning_efficiency}, clearly indicate that \abbrev~achieves a significant reduction in learning cost compared to Frozen. This is a desirable property, especially considering the rapid evolution of pre-trained models, as it allows us to efficiently retrain the meta-model and leverage the latest knowledge sources. Additionally, Tab.~\ref{tab:learning_efficiency} shows the inference cost of the representation obtained from frozen pre-trained models in bottom rows. Notably, our proposed architecture enables flexible selections for feature extractors, providing superior scalability and generalizability.

\begin{table}[t]
\renewcommand{\arraystretch}{1.0}
\footnotesize
\begin{center}
\begin{tabular}{l|ccc}
\toprule
\multirow{2}{*}{Setting} & VQAv2 & SNLI-VE & NLVR2 \\
& val & dev / test & dev / test \\
\midrule
 Default & 44.42 & 53.35/53.23  & 54.97/56.39 \\
\midrule
\multicolumn{4}{l}{\textit{Scale of meta-model}} \\[-0.2ex]
\midrule
124M $\rightarrow$ 82M 
    & 42.67 {\scriptsize \color{red} -1.75} 
    & 51.89/52.00 {\scriptsize \color{red} -1.35}
    & 53.25/54.44 {\scriptsize \color{red} -1.84} \\
\midrule
\multicolumn{4}{l}{\textit{Scale of pre-training dataset}} \\[-0.2ex]
\midrule
4M $\rightarrow$ 8M
    & 45.22 {\scriptsize \color{mygreen} +0.80}
    & 53.40/54.00 {\scriptsize \color{mygreen} +0.41}
    & 55.02/56.60 {\scriptsize \color{mygreen} +0.13} \\
\midrule
\multicolumn{4}{l}{\textit{Sources of data representations}} \\[-0.2ex]
\midrule
All $\rightarrow$ VL
    & 43.87 {\scriptsize \color{red} -0.55}
    & 52.89/53.00 {\scriptsize \color{red} -0.35}
    & 53.55/55.31 {\scriptsize \color{red} -1.25} \\
All $\rightarrow$ V + L
    & 36.14 {\scriptsize \color{red} -8.28}
    & 51.68/51.28 {\scriptsize \color{red} -1.81}
    & 52.90/52.67 {\scriptsize \color{red} -2.90} \\
\midrule
\multicolumn{4}{l}{\textit{Different VL Models}} \\[-0.2ex]
\midrule
\cite{Dou_2022_CVPR} $\rightarrow$~\cite{li2023blip} 
    & 41.78 {\scriptsize \color{red} -2.64}
    & 53.56/53.65 {\scriptsize \color{mygreen} +0.32}
    & 57.03/56.85 {\scriptsize \color{mygreen} +1.26} \\
\cite{Dou_2022_CVPR} $\rightarrow$~\cite{pmlr-v139-kim21k}
    & 37.46 {\scriptsize \color{red} -6.96}
    & 45.50/45.06 {\scriptsize \color{red} -8.01}
    & 52.84/52.94 {\scriptsize \color{red} -2.79} \\
\midrule
\multicolumn{4}{l}{\textit{Demonstration Selection}} \\[-0.2ex]
\midrule
w/ $\rightarrow$ w/o 
    & 42.80 {\scriptsize \color{red} -1.62}
    & 52.30/52.05 {\scriptsize \color{red} -1.12}
    & 52.46/53.39 {\scriptsize \color{red} -2.76} \\
\bottomrule
\end{tabular}
\end{center}
\caption{Different settings of \abbrev. Across settings, the in-demo ratio is 0.2 and the number of demonstrations is 4 for fair comparisons, which may not yield optimal values for tasks.}
\label{tab:different_settings}
\end{table}

\begin{table}[t]
\renewcommand{\arraystretch}{1.0}
\footnotesize
\begin{center}
\begin{tabular}{l|cc|cc}
\toprule
\multirow{2}{*}{Scale} & \multirow{2}{*}{\makecell{Input \\ Size}} & \multirow{2}{*}{\makecell{Learnable \\ Params.}} & \multirow{2}{*}{\makecell{Computation \\ (GFLOPs)}} & \multirow{2}{*}{\makecell{Memory \\ (GB)}} \\
& & & \\
\midrule
\multicolumn{5}{l}{\textit{Frozen~\cite{NEURIPS2021_01b7575c}}} \\[-0.2ex]
\midrule
Small & 1 & 86.39M & 1898.80 & 28.44 \\
Large & 1 & 304.35M & 2162.63 & 38.79 \\
\midrule
\multicolumn{5}{l}{\textit{\abbrev~(ours)}} \\[-0.2ex]
\midrule
Small & 1 & 81.91M & 2.15 & 2.18 \\
Large & 1 & 354.82M & 14.67 & 14.93 \\
\midrule
\ \ - VL Feats. & 9 & - & 301.06 & 3.26 \\ 
\ \ - V Feats. & 9 & - & 281.05 & 3.06 \\ 
\ \ - L Feats. & 9 & - & 792.95 & 11.45 \\ 
\bottomrule
\end{tabular}
\end{center}
\caption{Learning efficiency comparisons of \abbrev~with~\cite{NEURIPS2021_01b7575c}. The cost includes both forward and backward processes if required. We report the cost of feature inference with an input size of 9 for 8 demonstrations and 1 query data.}
\label{tab:learning_efficiency}
\end{table}

\section{Discussion and Conclusion}
Current ICL techniques in the vision-language (VL) domain are heavily reliant on large pre-trained language models, which may hinder their scalability and applicability. In this paper, we identify that this dependence arises from the ambiguous objective of acquiring ICL. To this end, we propose a novel framework, named SINC, that decouples the acquisition of ICL from VL pre-training and incentivizes it from both architectural and data perspectives. Our proposed method not only achieves superior performance compared to previous methods but also has a lower learning cost, making it advantageous for leveraging pre-trained models in a black-box setting. This property is particularly crucial for models with inaccessible parameters, such as ChatGPT. Moreover, SINC provides a general interface for exploring ICL in real-world applications and uncovering properties that can be utilized in future studies. We envision various directions for future research based on the proposed framework, e.g., alleviating the trade-off for learning with different prompts, enabling controllability conditioned on the given demonstrations, and facilitating generalization for higher shot numbers. We hope that our framework and the perspectives provided in this paper will further drive the development of ICL methods in the VL domain.

\section*{Acknowledgement}
\label{sec:ack} 
This work was supported in part by the National Science and Technology Council of Taiwan under Grants NSTC-109-2221-E-009-114-MY3 and NSTC-112-2221-E-A49-094-MY3.

{\small
\bibliographystyle{ieee_fullname}
\bibliography{main}

\begin{thebibliography}{10}\itemsep=-1pt

\bibitem{akyurek2022learning}
Ekin Aky{\"u}rek, Jacob Andreas, Dale Schuurmans, Tengyu Ma, and Denny Zhou.
\newblock What learning algorithm is in-context learning? investigations with
  linear models.
\newblock In {\em International Conference on Learning Representations}, pages
  1--8, 2023.

\bibitem{alayrac2022flamingo}
Jean-Baptiste Alayrac, Jeff Donahue, Pauline Luc, Antoine Miech, Iain Barr,
  Yana Hasson, Karel Lenc, Arthur Mensch, Katherine Millican, Malcolm Reynolds,
  Roman Ring, Eliza Rutherford, Serkan Cabi, Tengda Han, Zhitao Gong, Sina
  Samangooei, Marianne Monteiro, Jacob Menick, Sebastian Borgeaud, Andrew
  Brock, Aida Nematzadeh, Sahand Sharifzadeh, Mikolaj Binkowski, Ricardo
  Barreira, Oriol Vinyals, Andrew Zisserman, and Karen Simonyan.
\newblock Flamingo: a visual language model for few-shot learning.
\newblock In {\em Advances in Neural Information Processing Systems}, pages
  1--8, 2022.

\bibitem{anil2022exploring}
Cem Anil, Yuhuai Wu, Anders~Johan Andreassen, Aitor Lewkowycz, Vedant Misra,
  Vinay~Venkatesh Ramasesh, Ambrose Slone, Guy Gur-Ari, Ethan Dyer, and Behnam
  Neyshabur.
\newblock Exploring length generalization in large language models.
\newblock In {\em Advances in Neural Information Processing Systems}, pages
  1--8, 2022.

\bibitem{aribandi2022ext}
Vamsi Aribandi, Yi Tay, Tal Schuster, Jinfeng Rao, Huaixiu~Steven Zheng,
  Sanket~Vaibhav Mehta, Honglei Zhuang, Vinh~Q. Tran, Dara Bahri, Jianmo Ni,
  Jai Gupta, Kai Hui, Sebastian Ruder, and Donald Metzler.
\newblock Ext5: Towards extreme multi-task scaling for transfer learning.
\newblock In {\em International Conference on Learning Representations}, pages
  1--8, 2022.

\bibitem{bansal2022rethinking}
Hritik Bansal, Karthik Gopalakrishnan, Saket Dingliwal, Sravan Bodapati, Katrin
  Kirchhoff, and Dan Roth.
\newblock Rethinking the role of scale for in-context learning: An
  interpretability-based case study at 66 billion scale.
\newblock {\em arXiv preprint arXiv:2212.09095}, pages 1--8, 2022.

\bibitem{NEURIPS2020_1457c0d6}
Tom Brown, Benjamin Mann, Nick Ryder, Melanie Subbiah, Jared~D Kaplan, Prafulla
  Dhariwal, Arvind Neelakantan, Pranav Shyam, Girish Sastry, Amanda Askell,
  Sandhini Agarwal, Ariel Herbert-Voss, Gretchen Krueger, Tom Henighan, Rewon
  Child, Aditya Ramesh, Daniel Ziegler, Jeffrey Wu, Clemens Winter, Chris
  Hesse, Mark Chen, Eric Sigler, Mateusz Litwin, Scott Gray, Benjamin Chess,
  Jack Clark, Christopher Berner, Sam McCandlish, Alec Radford, Ilya Sutskever,
  and Dario Amodei.
\newblock Language models are few-shot learners.
\newblock In {\em Advances in Neural Information Processing Systems}, pages
  1877--1901, 2020.

\bibitem{chan2022data}
Stephanie~C.Y. Chan, Adam Santoro, Andrew~Kyle Lampinen, Jane~X Wang, Aaditya~K
  Singh, Pierre~Harvey Richemond, James McClelland, and Felix Hill.
\newblock Data distributional properties drive emergent in-context learning in
  transformers.
\newblock In {\em Advances in Neural Information Processing Systems}, pages
  1--8, 2022.

\bibitem{chen-etal-2022-improving}
Mingda Chen, Jingfei Du, Ramakanth Pasunuru, Todor Mihaylov, Srini Iyer,
  Veselin Stoyanov, and Zornitsa Kozareva.
\newblock Improving in-context few-shot learning via self-supervised training.
\newblock In {\em Proceedings of the 2022 Conference of the North American
  Chapter of the Association for Computational Linguistics: Human Language
  Technologies}, pages 3558--3573, 2022.

\bibitem{chen2015microsoft}
Xinlei Chen, Hao Fang, Tsung-Yi Lin, Ramakrishna Vedantam, Saurabh Gupta, Piotr
  Doll{\'a}r, and C~Lawrence Zitnick.
\newblock Microsoft coco captions: Data collection and evaluation server.
\newblock {\em arXiv preprint arXiv:1504.00325}, pages 1--8, 2015.

\bibitem{chen-etal-2022-meta}
Yanda Chen, Ruiqi Zhong, Sheng Zha, George Karypis, and He He.
\newblock Meta-learning via language model in-context tuning.
\newblock In {\em Proceedings of the 60th Annual Meeting of the Association for
  Computational Linguistics (Volume 1: Long Papers)}, pages 719--730, 2022.

\bibitem{Chen_Shuai_2021}
Yi-Syuan Chen and Hong-Han Shuai.
\newblock Meta-transfer learning for low-resource abstractive summarization.
\newblock {\em Proceedings of the AAAI Conference on Artificial Intelligence},
  pages 12692--12700, 2021.

\bibitem{9992078}
Yi-Syuan Chen, Yun-Zhu Song, and Hong-Han Shuai.
\newblock Spec: Summary preference decomposition for low-resource abstractive
  summarization.
\newblock {\em IEEE/ACM Transactions on Audio, Speech, and Language
  Processing}, pages 603--618, 2023.

\bibitem{pmlr-v139-cho21a}
Jaemin Cho, Jie Lei, Hao Tan, and Mohit Bansal.
\newblock Unifying vision-and-language tasks via text generation.
\newblock In {\em Proceedings of the 38th International Conference on Machine
  Learning}, pages 1931--1942, 2021.

\bibitem{dai2022can}
Damai Dai, Yutao Sun, Li Dong, Yaru Hao, Zhifang Sui, and Furu Wei.
\newblock Why can gpt learn in-context? language models secretly perform
  gradient descent as meta optimizers.
\newblock {\em arXiv preprint arXiv:2212.10559}, pages 1--8, 2022.

\bibitem{devlin-etal-2019-bert}
Jacob Devlin, Ming-Wei Chang, Kenton Lee, and Kristina Toutanova.
\newblock {BERT}: Pre-training of deep bidirectional transformers for language
  understanding.
\newblock In {\em Proceedings of the 2019 Conference of the North {A}merican
  Chapter of the Association for Computational Linguistics: Human Language
  Technologies, Volume 1 (Long and Short Papers)}, pages 4171--4186, 2019.

\bibitem{dong2022survey}
Qingxiu Dong, Lei Li, Damai Dai, Ce Zheng, Zhiyong Wu, Baobao Chang, Xu Sun,
  Jingjing Xu, and Zhifang Sui.
\newblock A survey for in-context learning.
\newblock {\em arXiv preprint arXiv:2301.00234}, pages 1--8, 2022.

\bibitem{dosovitskiy2021an}
Alexey Dosovitskiy, Lucas Beyer, Alexander Kolesnikov, Dirk Weissenborn,
  Xiaohua Zhai, Thomas Unterthiner, Mostafa Dehghani, Matthias Minderer, Georg
  Heigold, Sylvain Gelly, Jakob Uszkoreit, and Neil Houlsby.
\newblock An image is worth 16x16 words: Transformers for image recognition at
  scale.
\newblock In {\em International Conference on Learning Representations}, pages
  1--8, 2021.

\bibitem{Dou_2022_CVPR}
Zi-Yi Dou, Yichong Xu, Zhe Gan, Jianfeng Wang, Shuohang Wang, Lijuan Wang,
  Chenguang Zhu, Pengchuan Zhang, Lu Yuan, Nanyun Peng, Zicheng Liu, and
  Michael Zeng.
\newblock An empirical study of training end-to-end vision-and-language
  transformers.
\newblock In {\em Proceedings of the IEEE/CVF Conference on Computer Vision and
  Pattern Recognition (CVPR)}, pages 18166--18176, 2022.

\bibitem{garg2022what}
Shivam Garg, Dimitris Tsipras, Percy Liang, and Gregory Valiant.
\newblock What can transformers learn in-context? a case study of simple
  function classes.
\newblock In {\em Advances in Neural Information Processing Systems}, pages
  1--8, 2022.

\bibitem{glass-etal-2020-span}
Michael Glass, Alfio Gliozzo, Rishav Chakravarti, Anthony Ferritto, Lin Pan,
  G~P~Shrivatsa Bhargav, Dinesh Garg, and Avi Sil.
\newblock Span selection pre-training for question answering.
\newblock In {\em Proceedings of the 58th Annual Meeting of the Association for
  Computational Linguistics}, pages 2773--2782, 2020.

\bibitem{goyal-etal-2022-training}
Tanya Goyal, Jiacheng Xu, Junyi~Jessy Li, and Greg Durrett.
\newblock Training dynamics for text summarization models.
\newblock In {\em Findings of the Association for Computational Linguistics:
  ACL 2022}, pages 2061--2073, 2022.

\bibitem{Goyal_2017_CVPR}
Yash Goyal, Tejas Khot, Douglas Summers-Stay, Dhruv Batra, and Devi Parikh.
\newblock Making the v in vqa matter: Elevating the role of image understanding
  in visual question answering.
\newblock In {\em Proceedings of the IEEE Conference on Computer Vision and
  Pattern Recognition (CVPR)}, pages 1--8, 2017.

\bibitem{holtzman-etal-2021-surface}
Ari Holtzman, Peter West, Vered Shwartz, Yejin Choi, and Luke Zettlemoyer.
\newblock Surface form competition: Why the highest probability answer isn{'}t
  always right.
\newblock In {\em Proceedings of the 2021 Conference on Empirical Methods in
  Natural Language Processing}, pages 7038--7051, 2021.

\bibitem{Huang_2021_CVPR}
Zhicheng Huang, Zhaoyang Zeng, Yupan Huang, Bei Liu, Dongmei Fu, and Jianlong
  Fu.
\newblock Seeing out of the box: End-to-end pre-training for vision-language
  representation learning.
\newblock In {\em Proceedings of the IEEE/CVF Conference on Computer Vision and
  Pattern Recognition (CVPR)}, pages 12976--12985, 2021.

\bibitem{huang2020pixel}
Zhicheng Huang, Zhaoyang Zeng, Bei Liu, Dongmei Fu, and Jianlong Fu.
\newblock Pixel-bert: Aligning image pixels with text by deep multi-modal
  transformers.
\newblock {\em arXiv preprint arXiv:2004.00849}, pages 1--8, 2020.

\bibitem{Ji_2023_CVPR}
Yatai Ji, Rongcheng Tu, Jie Jiang, Weijie Kong, Chengfei Cai, Wenzhe Zhao,
  Hongfa Wang, Yujiu Yang, and Wei Liu.
\newblock Seeing what you miss: Vision-language pre-training with semantic
  completion learning.
\newblock In {\em Proceedings of the IEEE/CVF Conference on Computer Vision and
  Pattern Recognition (CVPR)}, pages 6789--6798, 2023.

\bibitem{jimenez-gutierrez-etal-2022-thinking}
Bernal Jimenez~Gutierrez, Nikolas McNeal, Clayton Washington, You Chen, Lang
  Li, Huan Sun, and Yu Su.
\newblock Thinking about {GPT}-3 in-context learning for biomedical {IE}? think
  again.
\newblock In {\em Findings of the Association for Computational Linguistics:
  EMNLP 2022}, pages 4497--4512, 2022.

\bibitem{jin-etal-2022-good}
Woojeong Jin, Yu Cheng, Yelong Shen, Weizhu Chen, and Xiang Ren.
\newblock A good prompt is worth millions of parameters: Low-resource
  prompt-based learning for vision-language models.
\newblock In {\em Proceedings of the 60th Annual Meeting of the Association for
  Computational Linguistics (Volume 1: Long Papers)}, pages 2763--2775, 2022.

\bibitem{johnson2019billion}
Jeff Johnson, Matthijs Douze, and Herv{\'e} J{\'e}gou.
\newblock Billion-scale similarity search with {GPUs}.
\newblock {\em IEEE Transactions on Big Data}, pages 535--547, 2019.

\bibitem{kim-etal-2021-changes}
Boseop Kim, HyoungSeok Kim, Sang-Woo Lee, Gichang Lee, Donghyun Kwak, Jeon
  Dong~Hyeon, Sunghyun Park, Sungju Kim, Seonhoon Kim, Dongpil Seo, Heungsub
  Lee, Minyoung Jeong, Sungjae Lee, Minsub Kim, Suk~Hyun Ko, Seokhun Kim,
  Taeyong Park, Jinuk Kim, Soyoung Kang, Na-Hyeon Ryu, Kang~Min Yoo, Minsuk
  Chang, Soobin Suh, Sookyo In, Jinseong Park, Kyungduk Kim, Hiun Kim, Jisu
  Jeong, Yong~Goo Yeo, Donghoon Ham, Dongju Park, Min~Young Lee, Jaewook Kang,
  Inho Kang, Jung-Woo Ha, Woomyoung Park, and Nako Sung.
\newblock What changes can large-scale language models bring? intensive study
  on {H}yper{CLOVA}: Billions-scale {K}orean generative pretrained
  transformers.
\newblock In {\em Proceedings of the 2021 Conference on Empirical Methods in
  Natural Language Processing}, pages 3405--3424, 2021.

\bibitem{pmlr-v139-kim21k}
Wonjae Kim, Bokyung Son, and Ildoo Kim.
\newblock Vilt: Vision-and-language transformer without convolution or region
  supervision.
\newblock In {\em Proceedings of the 38th International Conference on Machine
  Learning}, pages 5583--5594, 2021.

\bibitem{krishna2017visual}
Ranjay Krishna, Yuke Zhu, Oliver Groth, Justin Johnson, Kenji Hata, Joshua
  Kravitz, Stephanie Chen, Yannis Kalantidis, Li-Jia Li, David~A Shamma, et~al.
\newblock Visual genome: Connecting language and vision using crowdsourced
  dense image annotations.
\newblock {\em International journal of computer vision}, pages 32--73, 2017.

\bibitem{lambiotte2013burstiness}
Renaud Lambiotte, Lionel Tabourier, and Jean-Charles Delvenne.
\newblock Burstiness and spreading on temporal networks.
\newblock {\em The European Physical Journal B}, pages 1--4, 2013.

\bibitem{li2023blip}
Junnan Li, Dongxu Li, Silvio Savarese, and Steven Hoi.
\newblock Blip-2: Bootstrapping language-image pre-training with frozen image
  encoders and large language models.
\newblock {\em arXiv preprint arXiv:2301.12597}, pages 1--8, 2023.

\bibitem{NEURIPS2021_50525975}
Junnan Li, Ramprasaath Selvaraju, Akhilesh Gotmare, Shafiq Joty, Caiming Xiong,
  and Steven Chu~Hong Hoi.
\newblock Align before fuse: Vision and language representation learning with
  momentum distillation.
\newblock In {\em Advances in Neural Information Processing Systems}, pages
  9694--9705, 2021.

\bibitem{liang2022mvit}
Hanxue Liang, Zhiwen Fan, Rishov Sarkar, Ziyu Jiang, Tianlong Chen, Kai Zou, Yu
  Cheng, Cong Hao, and Zhangyang Wang.
\newblock M{\textthreesuperior}vit: Mixture-of-experts vision transformer for
  efficient multi-task learning with model-accelerator co-design.
\newblock In {\em Advances in Neural Information Processing Systems}, pages
  1--8, 2022.

\bibitem{liu-etal-2022-makes}
Jiachang Liu, Dinghan Shen, Yizhe Zhang, Bill Dolan, Lawrence Carin, and Weizhu
  Chen.
\newblock What makes good in-context examples for {GPT}-3?
\newblock In {\em Proceedings of Deep Learning Inside Out (DeeLIO 2022): The
  3rd Workshop on Knowledge Extraction and Integration for Deep Learning
  Architectures}, pages 100--114, 2022.

\bibitem{10.1145/3560815}
Pengfei Liu, Weizhe Yuan, Jinlan Fu, Zhengbao Jiang, Hiroaki Hayashi, and
  Graham Neubig.
\newblock Pre-train, prompt, and predict: A systematic survey of prompting
  methods in natural language processing.
\newblock {\em ACM Comput. Surv.}, pages 1--35, 2023.

\bibitem{liu2019roberta}
Yinhan Liu, Myle Ott, Naman Goyal, Jingfei Du, Mandar Joshi, Danqi Chen, Omer
  Levy, Mike Lewis, Luke Zettlemoyer, and Veselin Stoyanov.
\newblock Roberta: A robustly optimized bert pretraining approach.
\newblock {\em arXiv preprint arXiv:1907.11692}, pages 1--8, 2019.

\bibitem{lu-etal-2022-fantastically}
Yao Lu, Max Bartolo, Alastair Moore, Sebastian Riedel, and Pontus Stenetorp.
\newblock Fantastically ordered prompts and where to find them: Overcoming
  few-shot prompt order sensitivity.
\newblock In {\em Proceedings of the 60th Annual Meeting of the Association for
  Computational Linguistics (Volume 1: Long Papers)}, pages 8086--8098, 2022.

\bibitem{min-etal-2022-noisy}
Sewon Min, Mike Lewis, Hannaneh Hajishirzi, and Luke Zettlemoyer.
\newblock Noisy channel language model prompting for few-shot text
  classification.
\newblock In {\em Proceedings of the 60th Annual Meeting of the Association for
  Computational Linguistics (Volume 1: Long Papers)}, pages 5316--5330, 2022.

\bibitem{min-etal-2022-metaicl}
Sewon Min, Mike Lewis, Luke Zettlemoyer, and Hannaneh Hajishirzi.
\newblock {M}eta{ICL}: Learning to learn in context.
\newblock In {\em Proceedings of the 2022 Conference of the North American
  Chapter of the Association for Computational Linguistics: Human Language
  Technologies}, pages 2791--2809, 2022.

\bibitem{min2022rethinking}
Sewon Min, Xinxi Lyu, Ari Holtzman, Mikel Artetxe, Mike Lewis, Hannaneh
  Hajishirzi, and Luke Zettlemoyer.
\newblock Rethinking the role of demonstrations: What makes in-context learning
  work?
\newblock {\em arXiv preprint arXiv:2202.12837}, pages 1--8, 2022.

\bibitem{mishra-etal-2022-reframing}
Swaroop Mishra, Daniel Khashabi, Chitta Baral, Yejin Choi, and Hannaneh
  Hajishirzi.
\newblock Reframing instructional prompts to {GPT}k{'}s language.
\newblock In {\em Findings of the Association for Computational Linguistics:
  ACL 2022}, pages 589--612, 2022.

\bibitem{olsson2022context}
Catherine Olsson, Nelson Elhage, Neel Nanda, Nicholas Joseph, Nova DasSarma,
  Tom Henighan, Ben Mann, Amanda Askell, Yuntao Bai, Anna Chen, et~al.
\newblock In-context learning and induction heads.
\newblock {\em arXiv preprint arXiv:2209.11895}, pages 1--8, 2022.

\bibitem{NIPS2011_5dd9db5e}
Vicente Ordonez, Girish Kulkarni, and Tamara Berg.
\newblock Im2text: Describing images using 1 million captioned photographs.
\newblock In {\em Advances in Neural Information Processing Systems}, pages
  1--8, 2011.

\bibitem{pilault2021conditionally}
Jonathan Pilault, Amine~El hattami, and Christopher Pal.
\newblock Conditionally adaptive multi-task learning: Improving transfer
  learning in {NLP} using fewer parameters \& less data.
\newblock In {\em International Conference on Learning Representations}, pages
  1--8, 2021.

\bibitem{pmlr-v139-radford21a}
Alec Radford, Jong~Wook Kim, Chris Hallacy, Aditya Ramesh, Gabriel Goh,
  Sandhini Agarwal, Girish Sastry, Amanda Askell, Pamela Mishkin, Jack Clark,
  Gretchen Krueger, and Ilya Sutskever.
\newblock Learning transferable visual models from natural language
  supervision.
\newblock In {\em Proceedings of the 38th International Conference on Machine
  Learning}, pages 8748--8763, 2021.

\bibitem{radford2019language}
Alec Radford, Jeffrey Wu, Rewon Child, David Luan, Dario Amodei, Ilya
  Sutskever, et~al.
\newblock Language models are unsupervised multitask learners.
\newblock {\em OpenAI blog}, pages 1--8, 2019.

\bibitem{ram-etal-2021-shot}
Ori Ram, Yuval Kirstain, Jonathan Berant, Amir Globerson, and Omer Levy.
\newblock Few-shot question answering by pretraining span selection.
\newblock In {\em Proceedings of the 59th Annual Meeting of the Association for
  Computational Linguistics and the 11th International Joint Conference on
  Natural Language Processing (Volume 1: Long Papers)}, pages 3066--3079, 2021.

\bibitem{razeghi2022impact}
Yasaman Razeghi, Robert~L Logan~IV, Matt Gardner, and Sameer Singh.
\newblock Impact of pretraining term frequencies on few-shot numerical
  reasoning.
\newblock In {\em Findings of the Association for Computational Linguistics:
  EMNLP 2022}, pages 840--854, 2022.

\bibitem{rubin-etal-2022-learning}
Ohad Rubin, Jonathan Herzig, and Jonathan Berant.
\newblock Learning to retrieve prompts for in-context learning.
\newblock In {\em Proceedings of the 2022 Conference of the North American
  Chapter of the Association for Computational Linguistics: Human Language
  Technologies}, pages 2655--2671, 2022.

\bibitem{sennrich-etal-2016-neural}
Rico Sennrich, Barry Haddow, and Alexandra Birch.
\newblock Neural machine translation of rare words with subword units.
\newblock In {\em Proceedings of the 54th Annual Meeting of the Association for
  Computational Linguistics (Volume 1: Long Papers)}, pages 1715--1725, 2016.

\bibitem{serrano2009modeling}
M~{\'A}ngeles Serrano, Alessandro Flammini, and Filippo Menczer.
\newblock Modeling statistical properties of written text.
\newblock {\em PloS one}, page e5372, 2009.

\bibitem{sharma-etal-2018-conceptual}
Piyush Sharma, Nan Ding, Sebastian Goodman, and Radu Soricut.
\newblock Conceptual captions: A cleaned, hypernymed, image alt-text dataset
  for automatic image captioning.
\newblock In {\em Proceedings of the 56th Annual Meeting of the Association for
  Computational Linguistics (Volume 1: Long Papers)}, pages 2556--2565, 2018.

\bibitem{shin-etal-2022-effect}
Seongjin Shin, Sang-Woo Lee, Hwijeen Ahn, Sungdong Kim, HyoungSeok Kim, Boseop
  Kim, Kyunghyun Cho, Gichang Lee, Woomyoung Park, Jung-Woo Ha, and Nako Sung.
\newblock On the effect of pretraining corpora on in-context learning by a
  large-scale language model.
\newblock In {\em Proceedings of the 2022 Conference of the North American
  Chapter of the Association for Computational Linguistics: Human Language
  Technologies}, pages 5168--5186, 2022.

\bibitem{singla2022salient}
Sahil Singla and Soheil Feizi.
\newblock Salient imagenet: How to discover spurious features in deep learning?
\newblock In {\em International Conference on Learning Representations}, pages
  1--8, 2022.

\bibitem{song-etal-2022-clip}
Haoyu Song, Li Dong, Weinan Zhang, Ting Liu, and Furu Wei.
\newblock {CLIP} models are few-shot learners: Empirical studies on {VQA} and
  visual entailment.
\newblock In {\em Proceedings of the 60th Annual Meeting of the Association for
  Computational Linguistics (Volume 1: Long Papers)}, pages 6088--6100, 2022.

\bibitem{10.1145/3582688}
Yisheng Song, Ting Wang, Puyu Cai, Subrota~K Mondal, and Jyoti~Prakash Sahoo.
\newblock A comprehensive survey of few-shot learning: Evolution, applications,
  challenges, and opportunities.
\newblock {\em ACM Comput. Surv.}, pages 1--18, 2023.

\bibitem{song-etal-2022-improving}
Yun-Zhu Song, Yi-Syuan Chen, and Hong-Han Shuai.
\newblock Improving multi-document summarization through referenced flexible
  extraction with credit-awareness.
\newblock In {\em Proceedings of the 2022 Conference of the North American
  Chapter of the Association for Computational Linguistics: Human Language
  Technologies}, pages 1667--1681, 2022.

\bibitem{suhr-etal-2019-corpus}
Alane Suhr, Stephanie Zhou, Ally Zhang, Iris Zhang, Huajun Bai, and Yoav Artzi.
\newblock A corpus for reasoning about natural language grounded in
  photographs.
\newblock In {\em Proceedings of the 57th Annual Meeting of the Association for
  Computational Linguistics}, pages 6418--6428, 2019.

\bibitem{Sun_2019_CVPR}
Qianru Sun, Yaoyao Liu, Tat-Seng Chua, and Bernt Schiele.
\newblock Meta-transfer learning for few-shot learning.
\newblock In {\em Proceedings of the IEEE/CVF Conference on Computer Vision and
  Pattern Recognition (CVPR)}, pages 1--8, 2019.

\bibitem{NEURIPS2022_f8290ccc}
Yuchong Sun, Hongwei Xue, Ruihua Song, Bei Liu, Huan Yang, and Jianlong Fu.
\newblock Long-form video-language pre-training with multimodal temporal
  contrastive learning.
\newblock In {\em Advances in Neural Information Processing Systems}, pages
  38032--38045, 2022.

\bibitem{sung2022lst}
Yi-Lin Sung, Jaemin Cho, and Mohit Bansal.
\newblock {LST}: Ladder side-tuning for parameter and memory efficient transfer
  learning.
\newblock In {\em Advances in Neural Information Processing Systems}, pages
  1--8, 2022.

\bibitem{Sung_2022_CVPR}
Yi-Lin Sung, Jaemin Cho, and Mohit Bansal.
\newblock Vl-adapter: Parameter-efficient transfer learning for
  vision-and-language tasks.
\newblock In {\em Proceedings of the IEEE/CVF Conference on Computer Vision and
  Pattern Recognition (CVPR)}, pages 5227--5237, 2022.

\bibitem{pmlr-v139-triantafillou21a}
Eleni Triantafillou, Hugo Larochelle, Richard Zemel, and Vincent Dumoulin.
\newblock Learning a universal template for few-shot dataset generalization.
\newblock In {\em Proceedings of the 38th International Conference on Machine
  Learning}, pages 10424--10433, 2021.

\bibitem{NEURIPS2021_01b7575c}
Maria Tsimpoukelli, Jacob~L Menick, Serkan Cabi, S.~M.~Ali Eslami, Oriol
  Vinyals, and Felix Hill.
\newblock Multimodal few-shot learning with frozen language models.
\newblock In {\em Advances in Neural Information Processing Systems}, pages
  200--212, 2021.

\bibitem{NIPS2017_3f5ee243}
Ashish Vaswani, Noam Shazeer, Niki Parmar, Jakob Uszkoreit, Llion Jones,
  Aidan~N Gomez, \L~ukasz Kaiser, and Illia Polosukhin.
\newblock Attention is all you need.
\newblock In {\em Advances in Neural Information Processing Systems}, pages
  1--8, 2017.

\bibitem{von2022transformers}
Johannes von Oswald, Eyvind Niklasson, Ettore Randazzo, Jo{\~a}o Sacramento,
  Alexander Mordvintsev, Andrey Zhmoginov, and Max Vladymyrov.
\newblock Transformers learn in-context by gradient descent.
\newblock {\em arXiv preprint arXiv:2212.07677}, pages 1--8, 2022.

\bibitem{xie2019visual}
Ning Xie, Farley Lai, Derek Doran, and Asim Kadav.
\newblock Visual entailment: A novel task for fine-grained image understanding.
\newblock {\em arXiv preprint arXiv:1901.06706}, pages 1--8, 2019.

\bibitem{xu-lapata-2021-generating}
Yumo Xu and Mirella Lapata.
\newblock Generating query focused summaries from query-free resources.
\newblock In {\em Proceedings of the 59th Annual Meeting of the Association for
  Computational Linguistics and the 11th International Joint Conference on
  Natural Language Processing (Volume 1: Long Papers)}, pages 6096--6109, 2021.

\bibitem{Xue_2022_CVPR}
Hongwei Xue, Tiankai Hang, Yanhong Zeng, Yuchong Sun, Bei Liu, Huan Yang,
  Jianlong Fu, and Baining Guo.
\newblock Advancing high-resolution video-language representation with
  large-scale video transcriptions.
\newblock In {\em Proceedings of the IEEE/CVF Conference on Computer Vision and
  Pattern Recognition (CVPR)}, pages 5036--5045, 2022.

\bibitem{NEURIPS2021_23fa71cc}
Hongwei Xue, Yupan Huang, Bei Liu, Houwen Peng, Jianlong Fu, Houqiang Li, and
  Jiebo Luo.
\newblock Probing inter-modality: Visual parsing with self-attention for
  vision-and-language pre-training.
\newblock In {\em Advances in Neural Information Processing Systems}, pages
  4514--4528, 2021.

\bibitem{xue2023clipvip}
Hongwei Xue, Yuchong Sun, Bei Liu, Jianlong Fu, Ruihua Song, Houqiang Li, and
  Jiebo Luo.
\newblock {CLIP}-vip: Adapting pre-trained image-text model to video-language
  alignment.
\newblock In {\em The Eleventh International Conference on Learning
  Representations}, pages 1--8, 2023.

\bibitem{Yang_2022_CVPR}
Jinyu Yang, Jiali Duan, Son Tran, Yi Xu, Sampath Chanda, Liqun Chen, Belinda
  Zeng, Trishul Chilimbi, and Junzhou Huang.
\newblock Vision-language pre-training with triple contrastive learning.
\newblock In {\em Proceedings of the IEEE/CVF Conference on Computer Vision and
  Pattern Recognition (CVPR)}, pages 15671--15680, 2022.

\bibitem{yang2021empirical}
Zhengyuan Yang, Zhe Gan, Jianfeng Wang, Xiaowei Hu, Yumao Lu, Zicheng Liu, and
  Lijuan Wang.
\newblock An empirical study of gpt-3 for few-shot knowledge-based vqa.
\newblock In {\em AAAI}, pages 1--8, 2022.

\bibitem{kim2022ground}
Kang~Min Yoo, Junyeob Kim, Hyuhng~Joon Kim, Hyunsoo Cho, Hwiyeol Jo, Sang-Woo
  Lee, Sang-goo Lee, and Taeuk Kim.
\newblock Ground-truth labels matter: A deeper look into input-label
  demonstrations.
\newblock In {\em Proceedings of the 2022 Conference on Empirical Methods in
  Natural Language Processing}, pages 2422--2437, 2022.

\bibitem{NEURIPS2020_c8512d14}
Manzil Zaheer, Guru Guruganesh, Kumar~Avinava Dubey, Joshua Ainslie, Chris
  Alberti, Santiago Ontanon, Philip Pham, Anirudh Ravula, Qifan Wang, Li Yang,
  and Amr Ahmed.
\newblock Big bird: Transformers for longer sequences.
\newblock In {\em Advances in Neural Information Processing Systems}, pages
  17283--17297, 2020.

\bibitem{zhang-etal-2022-active}
Yiming Zhang, Shi Feng, and Chenhao Tan.
\newblock Active example selection for in-context learning.
\newblock In {\em Proceedings of the 2022 Conference on Empirical Methods in
  Natural Language Processing}, pages 9134--9148, 2022.

\bibitem{zhang-etal-2023-hyperpelt}
Zhengkun Zhang, Wenya Guo, Xiaojun Meng, Yasheng Wang, Yadao Wang, Xin Jiang,
  Qun Liu, and Zhenglu Yang.
\newblock {H}yper{PELT}: Unified parameter-efficient language model tuning for
  both language and vision-and-language tasks.
\newblock In {\em Findings of the Association for Computational Linguistics:
  ACL 2023}, pages 11442--11453, 2023.

\bibitem{pmlr-v139-zhao21c}
Zihao Zhao, Eric Wallace, Shi Feng, Dan Klein, and Sameer Singh.
\newblock Calibrate before use: Improving few-shot performance of language
  models.
\newblock In {\em Proceedings of the 38th International Conference on Machine
  Learning}, pages 12697--12706, 2021.

\bibitem{Zheng_2017_ICCV}
Heliang Zheng, Jianlong Fu, Tao Mei, and Jiebo Luo.
\newblock Learning multi-attention convolutional neural network for
  fine-grained image recognition.
\newblock In {\em Proceedings of the IEEE International Conference on Computer
  Vision (ICCV)}, 2017.

\end{thebibliography}
}

\end{document}